%% file: main.tex
\documentclass{article}
\usepackage[utf8]{inputenc}
\usepackage[margin=1.25in]{geometry}

\PassOptionsToPackage{sort}{natbib}
\PassOptionsToPackage{square}{natbib}
\usepackage{natbib}

\newif\ifacl

\usepackage{amsfonts}
\usepackage{amsthm}
\usepackage{amsmath}
\usepackage{amssymb}
\usepackage{xfrac} 

\usepackage{hyperref}
\hypersetup{
    colorlinks=true,
    linkcolor=blue,
    filecolor=blue, 
    urlcolor=magenta,
    citecolor=blue,
    }

\PassOptionsToPackage{usenames,dvipsnames}{xcolor}
\usepackage[capitalise]{cleveref}
\renewcommand{\Cref}{\cref}
\Crefname{section}{Sec.}{Secs.}
\Crefname{appendix}{App.}{Apps.}
\Crefname{algorithm}{Alg.}{Algs.}
\Crefname{table}{Tab.}{Tabs.}
\Crefname{theorem}{Thm.}{Thms.}

\usepackage[noend]{algorithmic}
\usepackage{algorithm}
\usepackage{multicol}
\usepackage{float}
\usepackage{multirow}

\usepackage{graphicx}
\graphicspath{./images/}
\usepackage[skip=0pt]{subcaption}

\usepackage[inline]{enumitem}
\setenumerate[1]{itemsep=0pt,partopsep=0pt,parsep=\parskip,topsep=5pt}
\setitemize[1]{itemsep=0pt,partopsep=0pt,parsep=\parskip,topsep=5pt}
\setdescription{itemsep=0pt,partopsep=0pt,parsep=\parskip,topsep=5pt}

\usepackage{xcolor}

\definecolor{darkgreen}{rgb}{0,0.4,0.0}

\makeatletter
\@ifundefined{theorem}{%
\newtheorem {theorem}{Theorem}}{}
\@ifundefined{lemma}{%
}{}
\@ifundefined{corollary}{%
}{}
\@ifundefined{conjecture}{%
}{}

\newcommand{\tree}{\mathcal{T}}
\newcommand{\psum}{\text{PrivateSum}}
\newcommand{\restart}{\mathcal{R}}

\usepackage[framemethod=tikz]{mdframed}

\usepackage[affil-sl]{authblk}

\makeatletter
\renewcommand\AB@affilsepx{,\quad \protect\Affilfont}
\makeatother

\title{Federated Learning of Gboard Language Models with Differential Privacy}
\author{Zheng Xu\thanks{Equal contribution, alphabetical order. Correspondence to \texttt{\{xuzheng,zhangyx\}@google.com}.}}
\author{Yanxiang Zhang$^*$} 
\author{Galen Andrew} 
\author{Christopher A. Choquette-Choo} 
\author{Peter Kairouz}
\author{H. Brendan McMahan}
\author{Jesse Rosenstock}
\author{Yuanbo Zhang}
\affil{Google}
\date{}
\begin{document}
\setboolean{acl}{false} 
\maketitle

\input{sec_abs}

\input{sec_intro}

\input{sec_method}

\input{sec_exp}

\input{sec_con}

\input{sec_ack}

\bibliographystyle{plainnat}
\bibliography{refer}

\appendix
\input{sec_app_dpg}

\end{document}

%% file: sec_abs.tex
\begin{abstract}
We train language models (LMs) with federated learning (FL) and differential privacy (DP) in the Google Keyboard (Gboard). We apply the DP-Follow-the-Regularized-Leader (DP-FTRL)~\citep{kairouz21b} algorithm to achieve meaningfully formal DP guarantees without requiring uniform sampling of client devices. 
To provide favorable privacy-utility trade-offs, we introduce a new client participation criterion and discuss the implication of its configuration in large scale systems. We show how quantile-based clip estimation~\citep{andrew2019differentially} can be combined with DP-FTRL to adaptively choose the clip norm during training or reduce the hyperparameter tuning in preparation for training. 
With the help of pretraining on public data, we train and deploy more than twenty Gboard LMs that achieve high utility and $\rho-$zCDP privacy guarantees with $\rho \in (0.2, 2)$, with two models additionally trained with secure aggregation~\citep{bonawitz2017practical}.
We are happy to announce that all the next word prediction neural network LMs in Gboard now have DP guarantees, and all future launches of Gboard neural network LMs will require DP guarantees. 
We summarize our experience and provide concrete suggestions on DP training for practitioners. 
\end{abstract}

%% file: sec_intro.tex

\section{Introduction}

\textbf{FL and Gboard LMs.}
In cross-device federated learning (FL), client devices collaboratively train a model without directly exchanging their local data \citep{kairouz2019advances}. Google Keyboard (Gboard) was an early adopter of FL to train models that improve the user experience, following data minimization principles \citep{bonawitz2021federated} to protect users' privacy from some risks. Language models (LMs) are trained with FL to support various features in Gboard, including Next Word Prediction (NWP), Smart Compose (SC), and  On-The-Fly rescoring (OTF). As illustrated in \cref{fig:features}, NWP \citep{hard2018gboard} uses an LM to suggest a word, which is triggered after a previous word is committed; SC provides longer inline suggestions to accelerate typing, which can be triggered per character when the confidence is high; OTF is used to re-rank the candidate words generated during typing before a word is committed. 

\ifthenelse{\boolean{acl}}{
\begin{figure}[t]
    \centering
    \includegraphics[width=\linewidth]{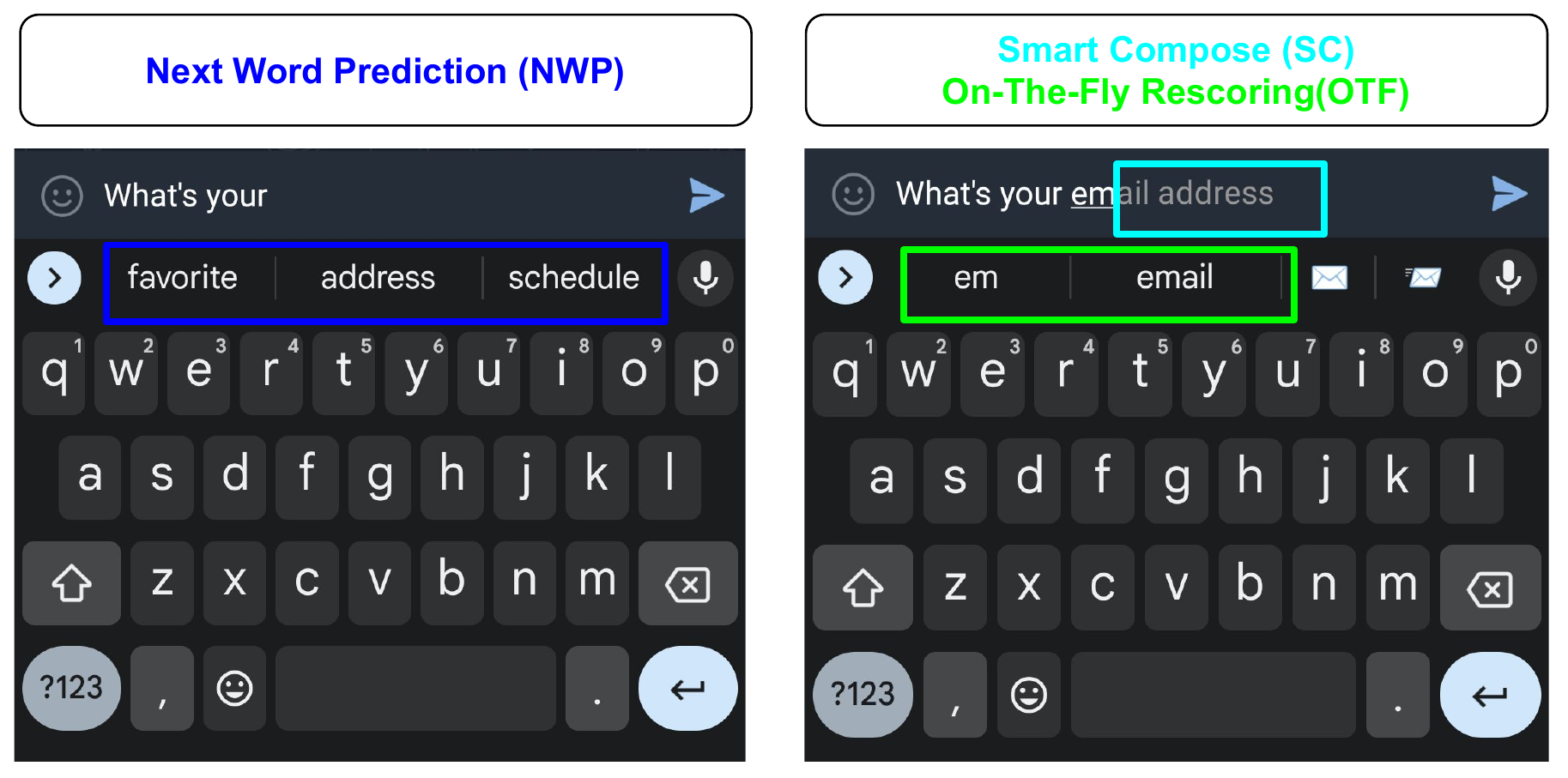}
    \caption{Gboard features supported by language models: NWP for next word, SC for inline suggestion, and OTF for candidates re-ranking.}
    \label{fig:features}
    \vspace{-0.6cm}
\end{figure}
\begin{figure*}[tbh]
    \centering
    \includegraphics[width=0.8\linewidth]{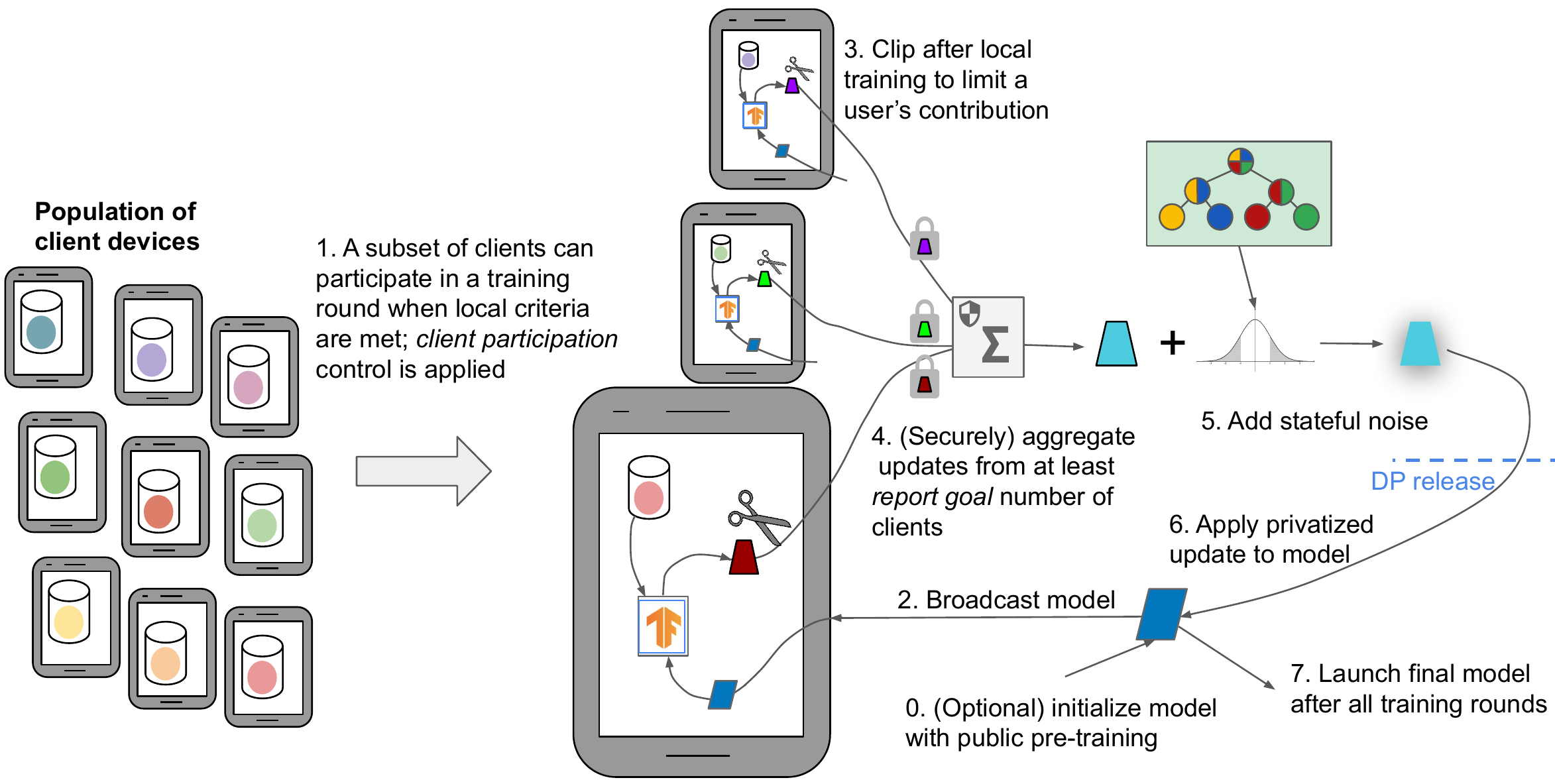}
    \caption{System overview of federated learning of Gboard language models with differential privacy and secure aggregation.}
    \label{fig:teaser}
    \vspace{-0.6cm}
\end{figure*}
}{
\begin{figure}[t]
    \centering
    \includegraphics[width=0.8\linewidth]{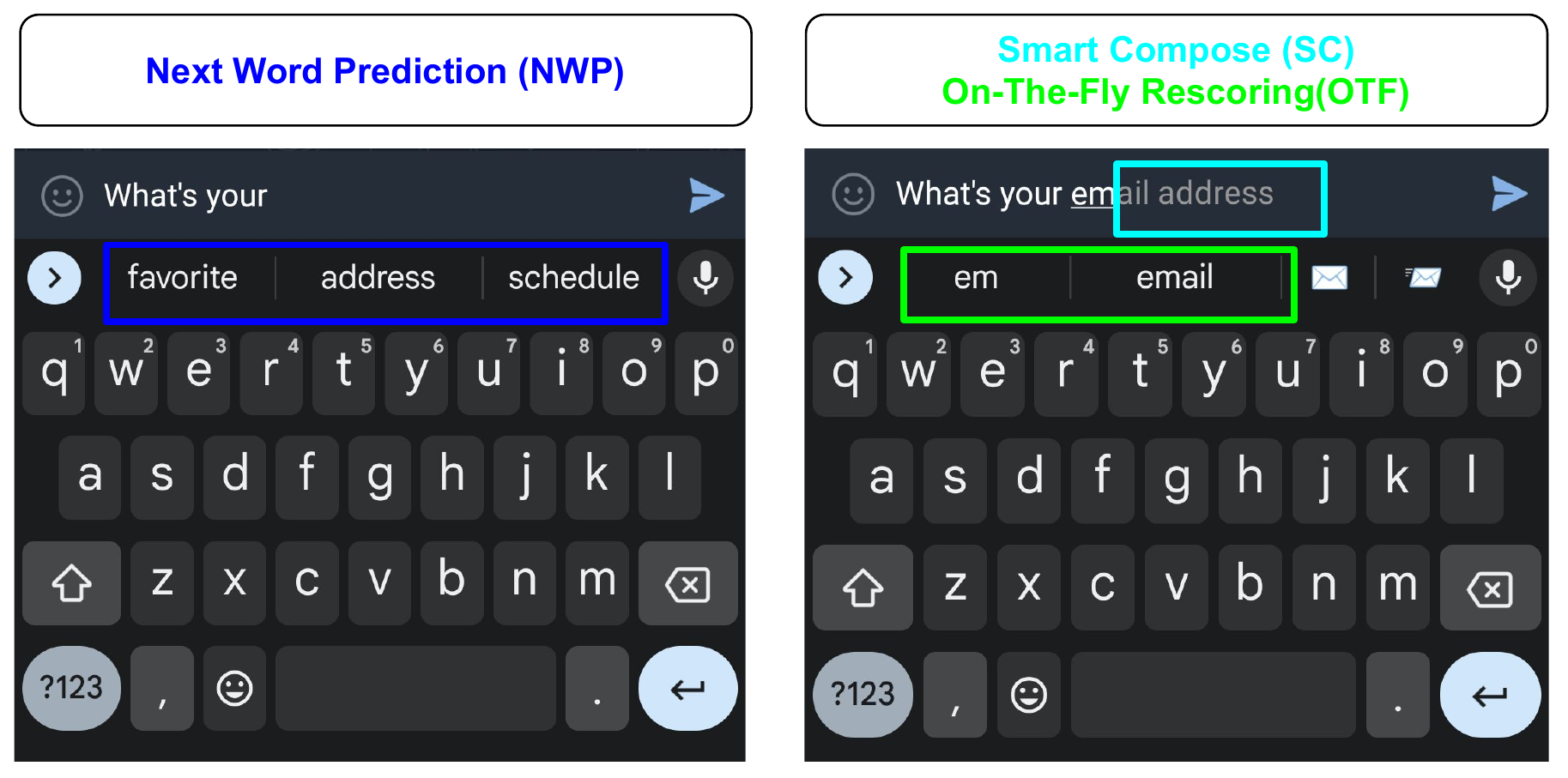}
    \caption{Gboard features supported by language models: NWP for next word, SC for inline suggestion, and OTF for candidates re-ranking.}
    \label{fig:features}
\end{figure}

\begin{figure*}[tbh]
    \centering
    \includegraphics[width=\linewidth]{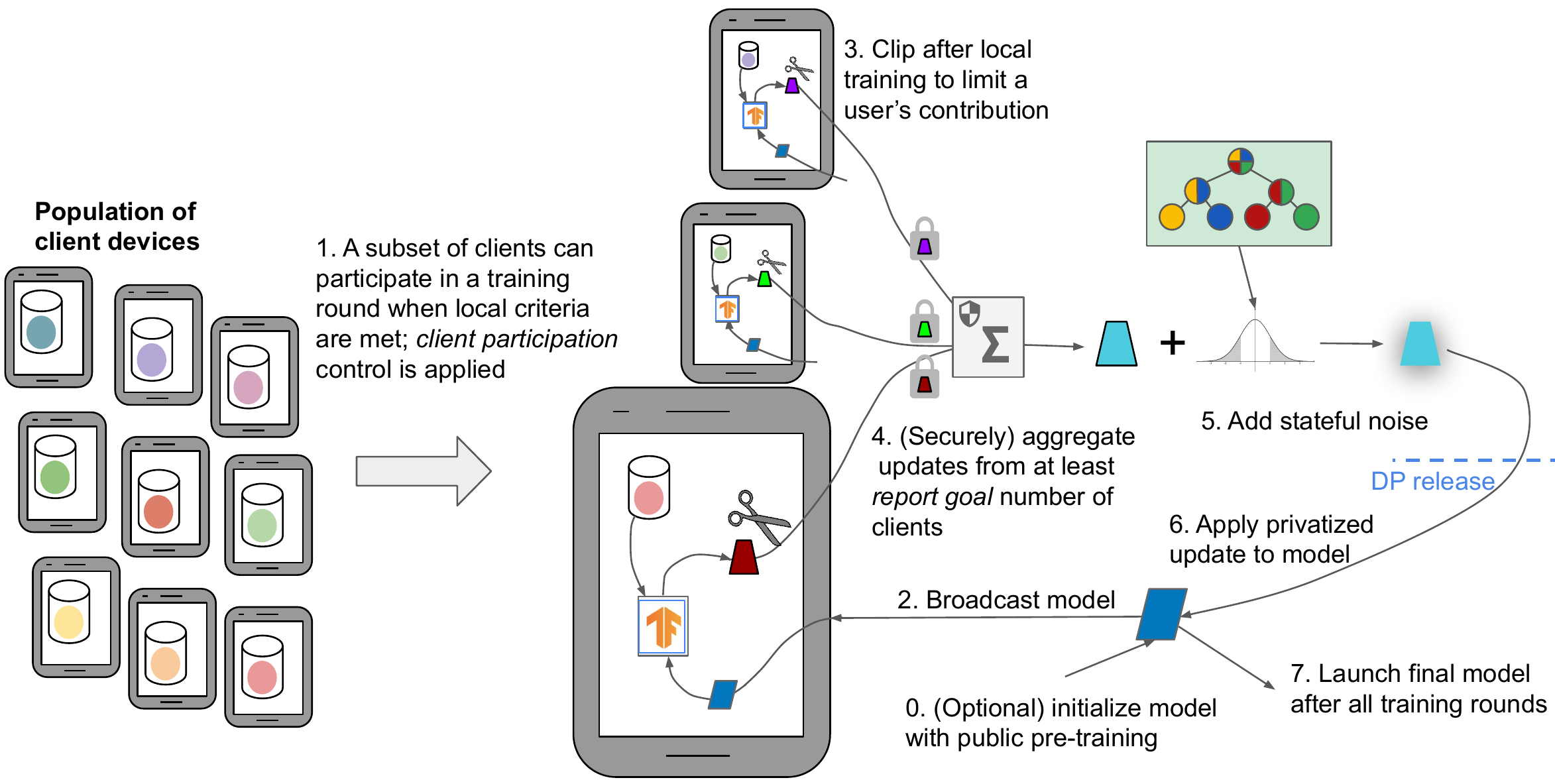}
    \caption{System overview of federated learning with differential privacy and secure aggregation.}
    \label{fig:teaser}
\end{figure*}
}

\paragraph{Models, metrics and tasks.} We train LMs with the same neural network (NN) architecture described in \citep{hard2018gboard}: a one-layer LSTM/CIFG of 670 hidden neurons, with input and output word-based embeddings of dimension 96. OTF LMs use a larger vocabulary ($\sim$ 30K words) compared to NWP LMs ($\sim$ 10--20K words); the number of parameters for models with a 10K/20K/30K vocabulary is 2.4M/4.4M/6.4M, respectively. SC is a downstream task that reuses NWP LMs without any retraining from data. We train NWP LMs and OTF LMs from populations of devices categorized by language and location. For example, en-US NWP denotes the task of training NWP model on data generated by devices using English in the United States.

Federated Averaging (FedAvg) \citep{mcmahan2017fedavg} and variants \citep{wang2021fieldguide} are popular FL training algorithms in practice. In each communication \emph{round}, the server will orchestrate a small subset of client devices for training and aggregate the resulting model deltas to update the global model. In a successful round, the system guarantees the number of clients participating in training is at least as large as the configured \emph{report goal} \citep{bonawitz2019towards}. A model is typically tested and deployed after training for several thousands of rounds. Top-1 in-vocab accuracy is used to track the utility during training and additional metrics for A/B testing are introduced in \cref{sec:exp}.

\paragraph{DP and DP-FTRL.} 
Differential privacy (DP) can be combined with FL to provide a formal guarantee that the trained model will not memorize specific users' data, which provides stronger privacy protection by executing data anonymization principles \citep{bonawitz2021federated,wang2021fieldguide}. \citet{ramaswamy2020training} applied DP-FedAvg~\citep{mcmahan18learning,geyer2017differentially}, a variant of DP-SGD~\citep{abadi2016deep} for user/client-level DP, to train production LMs in FL. \citet{ramaswamy2020training} demonstrated anonymization via empirical auditing techniques by \citet{carlini2019secret} but did not provide a formal DP guarantee.
Achieving a strong formal DP guarantee for DP-FedAvg would require privacy amplification-by-sampling, which necessitates sampling clients uniformly at random on each round. However, a cross-device FL system has limited control over client sampling as devices have to satisfy local criteria such as being charging and connected to an unmetered network to be eligible for participation \citep{bonawitz2019towards,balle2020privacy}. In contrast, we deploy a recent algorithm, DP-FTRL~\citep{kairouz21b}, allowing us to achieve strong privacy and utility for production models without uniform sampling assumptions.

\paragraph{Contributions.} We discuss our strategy and experience of training Gboard LMs with FL and DP. We introduce an algorithm that enables adaptive clipping \citep{andrew2019differentially} in DP-FTRL \citep{kairouz21b} (\cref{sec:adpftrl}), which can reliably estimate the clip norm to reduce hyperparameter tuning. We discuss the impact of scaling up computation and limiting client participation (\cref{sec:hyper}), and identify the algorithm and system configurations for the regime of strong privacy and utility. We also successfully apply pre-training (\cref{sec:pretrain}) to improve privacy and utility, which is (to the best of our knowledge) the first time pretraining is applied to training a DP model directly from users' data. 

We combine DP-FTRL with secure aggregation (SecAgg) to further strengthen the data minimization properties of our approach (\cref{sec:secagg}). \Cref{fig:teaser} provides a system overview of the techniques for training Gboard language models with federated learning and differential privacy. 
Finally, we summarize concrete suggestions for practitioners training differentially private models to deploy in production in (\cref{sec:practices}), and present and analyze twenty Gboard LMs trained with formal DP guarantees (\cref{sec:exp}). We are happy to announce that all the next word prediction neural network LMs in Gboard now have DP guarantees, and all future launches of Gboard neural network LMs will require DP guarantees. 

%% file: sec_method.tex
\input{sec_alg}
\section{DP FL in Practice} \label{sec:method}
\subsection{DP-FTRL and adaptive clipping} \label{sec:adpftrl}
\ifthenelse{\boolean{acl}}{
\begin{figure}[tbh]
\centering
\begin{subfigure}[b]{0.505\linewidth}
\centering
\includegraphics[width=\textwidth]{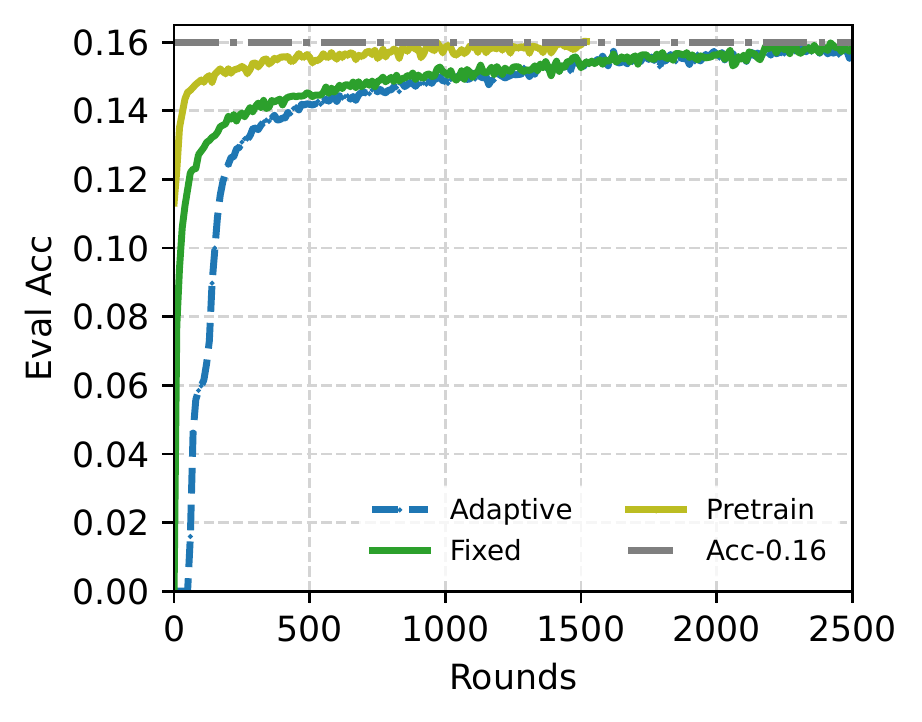}
\caption{Evaluation accuracy}
\label{fig:engb_acc}
\end{subfigure}
\begin{subfigure}[b]{0.48\linewidth}
\centering
\includegraphics[width=\textwidth]{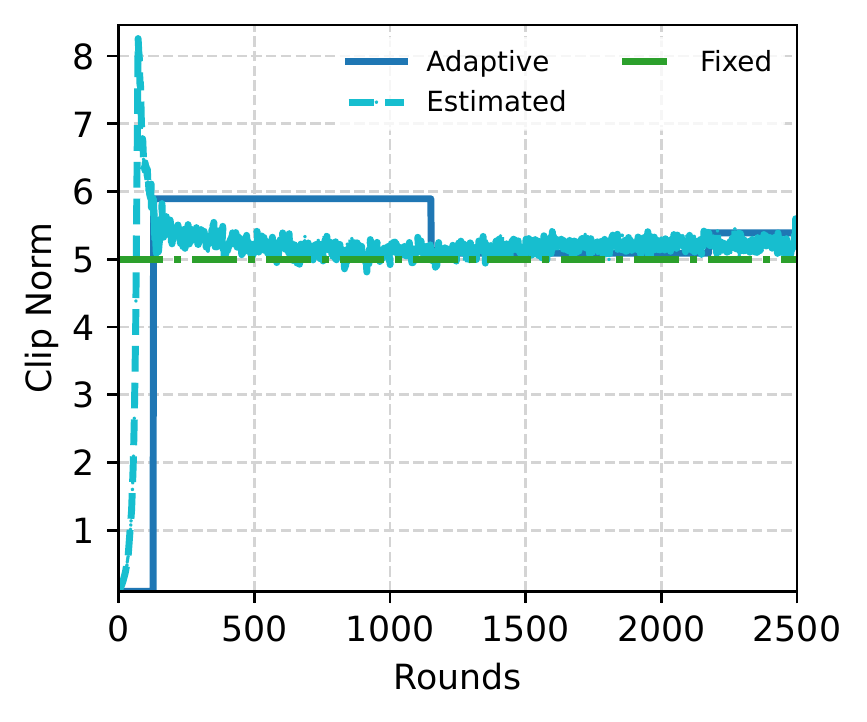}
\caption{Clip norm}
\label{fig:engb_clip}
\end{subfigure}
\caption{DP training of the en-GB NWP model. Adaptive clipping performs similar to fixed clipping, while achieves slightly weaker guarantees. Pre-training significantly reduces the number of rounds to reach the utility target, and achieves stronger guarantees.}\label{fig:engb}
\vspace{-0.5cm}
\end{figure}
}{
\begin{figure}[tbh]
\centering
\begin{subfigure}[b]{0.455\linewidth}
\centering
\includegraphics[width=\textwidth]{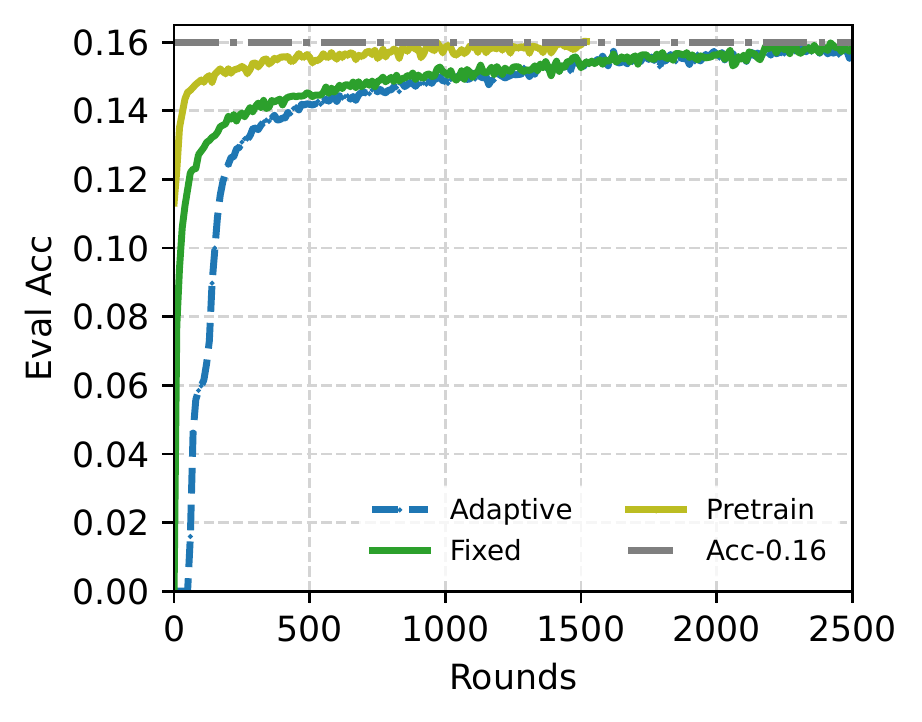}
\caption{Evaluation accuracy}
\label{fig:engb_acc}
\end{subfigure}
\begin{subfigure}[b]{0.43\linewidth}
\centering
\includegraphics[width=\textwidth]{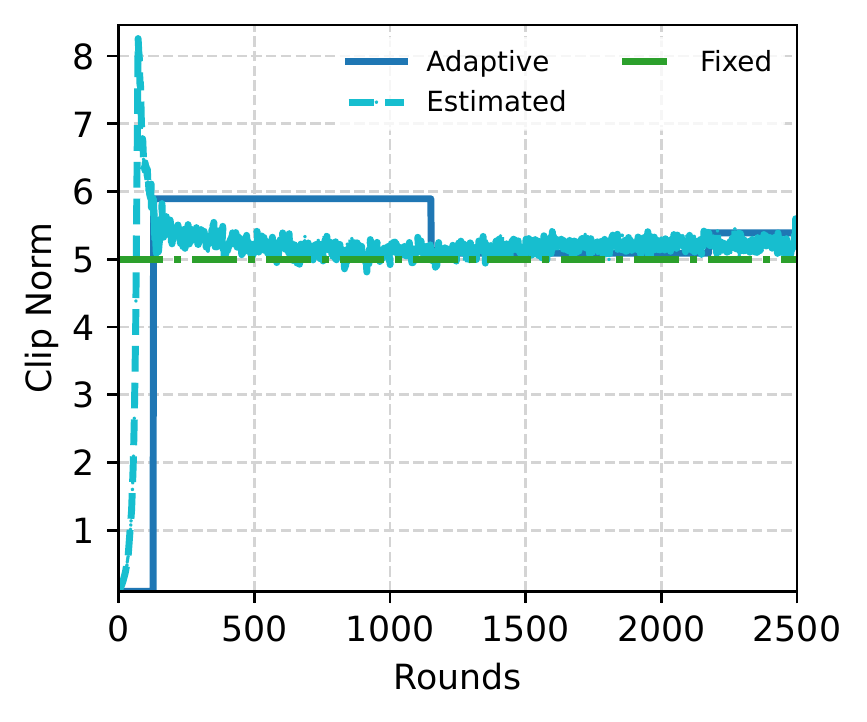}
\caption{Clip norm}
\label{fig:engb_clip}
\end{subfigure}
\caption{DP training of the en-GB NWP model. Adaptive clipping performs similar to fixed clipping, while achieves slightly weaker guarantees. Pre-training significantly reduces the number of rounds to reach the utility target, and achieves stronger guarantees.}\label{fig:engb}
\end{figure}
}

As described in \cref{algo:ada-ftrl}, we apply DP-FTRL in FL by modifying the FedAvg algorithm: clip the model update $\Delta$, and add noise when updating the global model. Two additional hyperparameters are introduced for DP: the \emph{clip norm $C$}, which bounds the norm of $\Delta$, and the \emph{noise multiplier $z$}, which determines the standard deviation $zC$ for the added Gaussian noise. We discuss clip norm in this section and defer the discussion of noise multiplier and other privacy related hyperparameters to \cref{sec:hyper}.

\citet{andrew2019differentially} introduced an adaptive clipping method that automatically adjusts the clip norm each round by privately estimating the norm of the model delta at a targeted quantile. However, adaptive clipping cannot be directly applied to DP-FTRL as the tree-based noise addition in DP-FTRL assumes a fixed clip norm across rounds. 
We integrate adaptive clipping in DP-FTRL through restarts, where the quantile estimate $C^t$ is continually tracked but only 
becomes an active clip norm $C_\theta$ upon tree restarting. As both the aggregated model delta $\tilde{\Delta}^t$ and the quantile $\tilde{b}^t$ use tree-based noise, we can directly use the privacy accounting in \citep{kairouz21b} by applying the noise transformation in \Cref{thm:equivalent_noise} 
\ifthenelse{\boolean{acl}}{in \cref{sec:ada_acc}}{}.

\ifthenelse{\boolean{acl}}{}{
\begin{theorem}[Privacy Accounting for Adaptive Clipping \citep{andrew2019differentially}]
\label{thm:equivalent_noise}
One step of DP-FTRL with adaptive clipping using $\sigma_b$ noise standard deviation on the clipped counts $\sum b_i^t$ and $z_\Delta$ noise multiplier on the vector sums $\sum \Delta_i^t$ is equivalent to one step of non-adaptive DP-FTRL with noise multiplier $z$ if we set $z_{\Delta} = \left( z^{-2} - (2\sigma_b)^{-2} \right)^{-\sfrac{1}{2}}$.
\end{theorem}
}

In practice, \Cref{algo:ada-ftrl} slightly inflates the noise for the model from $zC$ to $z_\Delta C$ and requires \emph{restarts} that complicate the privacy accounting for DP-FTRL. 
Moreover, we find that a fixed clip norm can achieve comparable or slightly better model utility, and is more robust in experiments with large report goal. For example, adaptive clipping for the de-DE NWP model experiences catastrophic failure and makes no progress in the first 1000 rounds. 

Nevertheless, adaptive clipping can reduce hyperparameter tuning for many tasks when privacy budget allows. \Cref{fig:engb} shows the evaluation accuracy and corresponding clip norm for DP training the en-GB NWP model with report goal 6500 and noise multiplier 7. The adaptive clip curve starts from a small initial clip norm to avoid catastrophic failure due to large initial noise and eventually catches up on accuracy. The estimated clip norm (quantile $\gamma=0.5$) stabilizes and we can fix the clip norm to $5$ based on the estimated value. The clip norm is relatively insensitive, especially when tuning together with the server learning rate. However, clip norm can have a wide tuning range across tasks and models, and quantile-based estimation is still useful for estimating a clip norm to be fixed. 

\subsection{DP parameters and system configuration} \label{sec:hyper} 
The privacy guarantees of DP-FTRL~\citep{kairouz21b} are affected by several factors: noise multiplier $z$, number of total rounds $T$, max participation (MaxP) of a client, and min separation (MinS) of rounds between the participation of the same client. The noise multiplier is a conventional parameter for controlling privacy-utility trade-off: large noise achieves strong privacy guarantees but can potentially hurt the utility. Achieving the same utility with smaller rounds $T$ can significantly improve the privacy guarantees. 
Next, we discuss the effect of MaxP and MinS, and the privacy-utility-computation trade-off for system configuration. 

\ifthenelse{\boolean{acl}}{
\begin{figure*}[thb]
\centering
\begin{subfigure}[b]{0.27\linewidth}
\centering
\includegraphics[width=\textwidth]{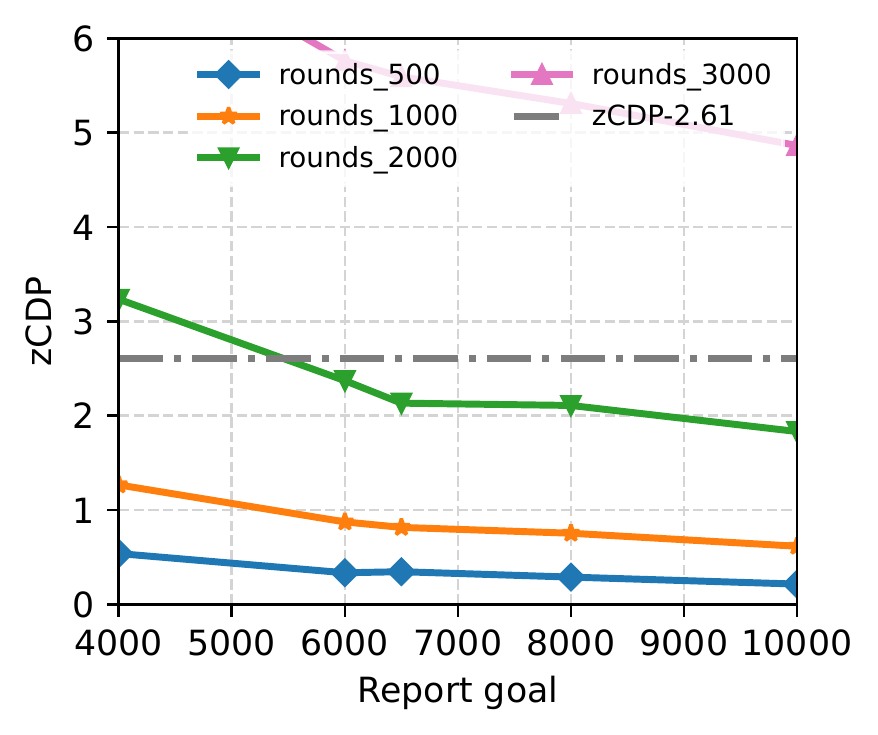}
\caption{Population 1M}
\label{fig:pop1m}
\end{subfigure}
\begin{subfigure}[b]{0.28\linewidth}
\centering
\includegraphics[width=\textwidth]{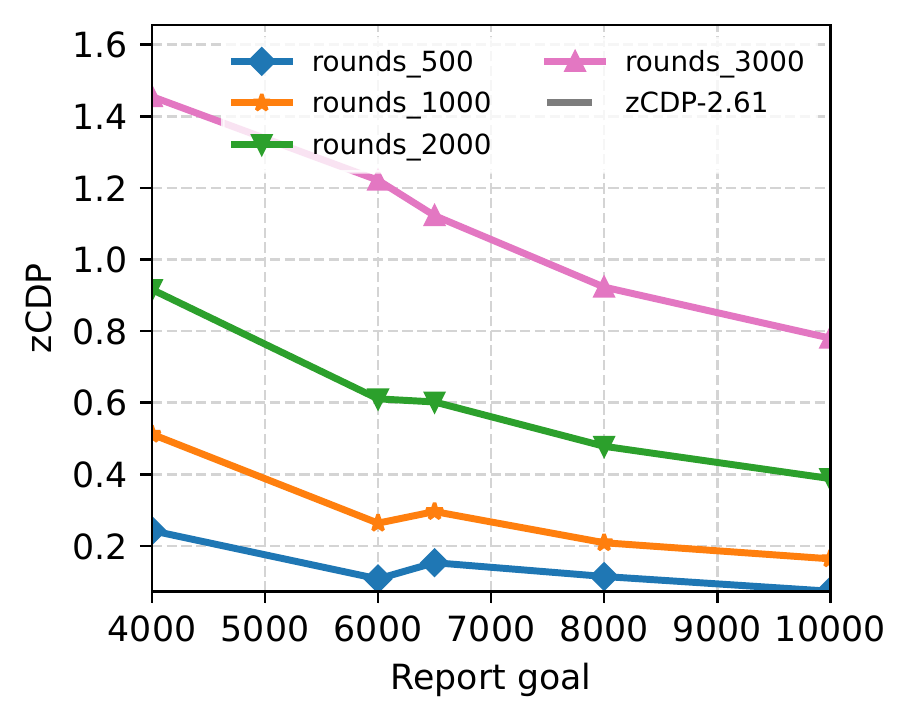}
\caption{Population 3M}
\label{fig:pop3m}
\end{subfigure}
\begin{subfigure}[b]{0.28\linewidth}
\centering
\includegraphics[width=\textwidth]{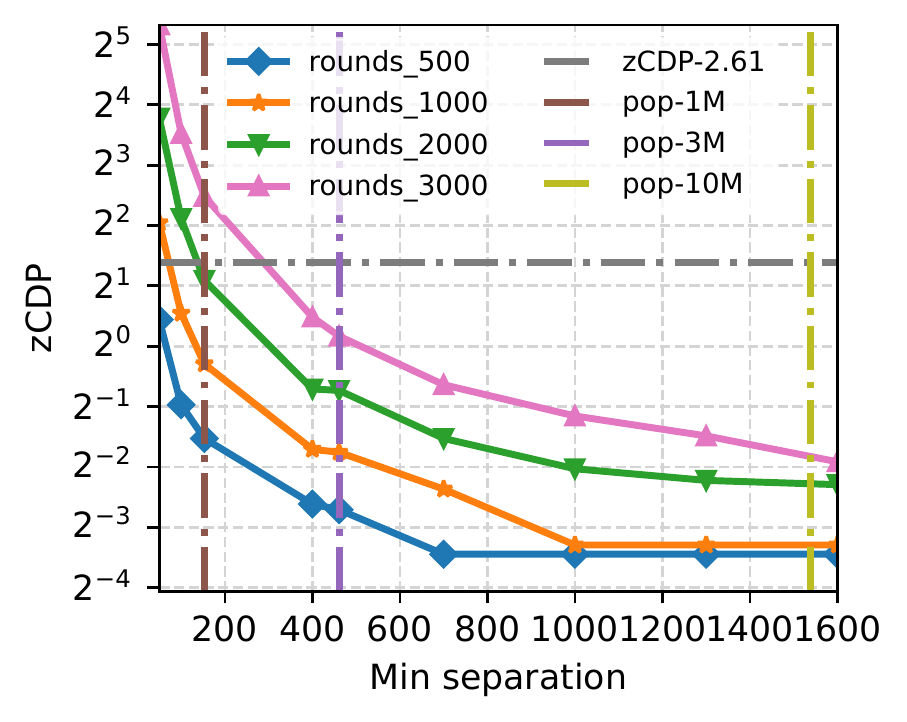}
\caption{Report goal 6500}
\label{fig:min_sep}
\end{subfigure}
\caption{The effect of population size, number of rounds, report goals, and min separation on DP-FTRL privacy guarantees. For a fixed number of rounds to achieve utility target, increasing report goal and min separation can achieve stronger guarantees measured by smaller zCDP. 
}\label{fig:privacy_computation}
\vspace{-0.5cm}
\end{figure*}
}{
\begin{figure*}[thb]
\centering
\begin{subfigure}[b]{0.325\linewidth}
\centering
\includegraphics[width=\textwidth]{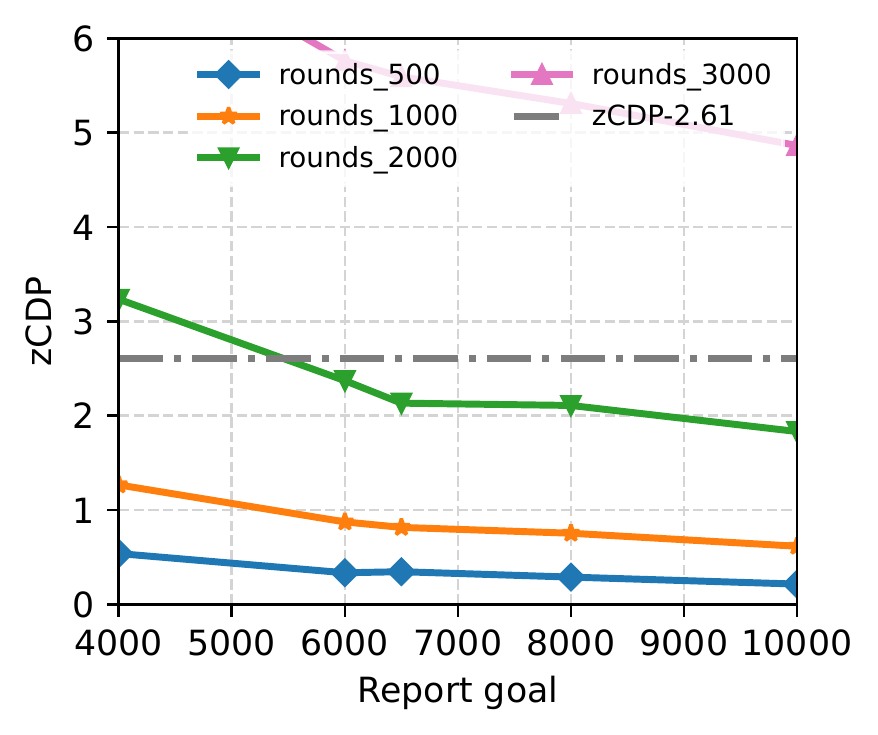}
\caption{Population 1M}
\label{fig:pop1m}
\end{subfigure}
\begin{subfigure}[b]{0.33\linewidth}
\centering
\includegraphics[width=\textwidth]{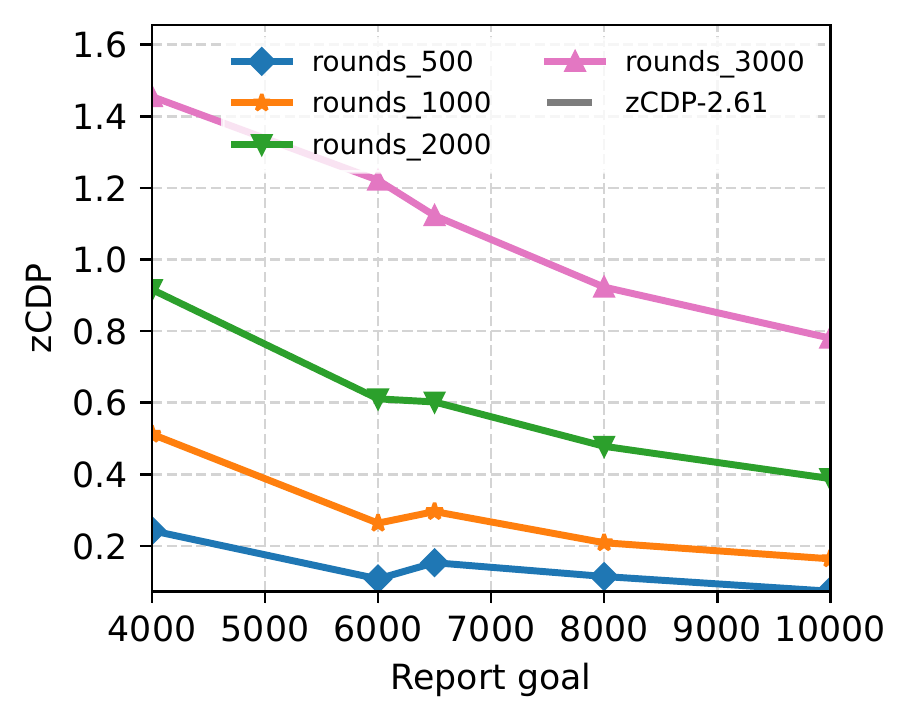}
\caption{Population 3M}
\label{fig:pop3m}
\end{subfigure}
\begin{subfigure}[b]{0.33\linewidth}
\centering
\includegraphics[width=\textwidth]{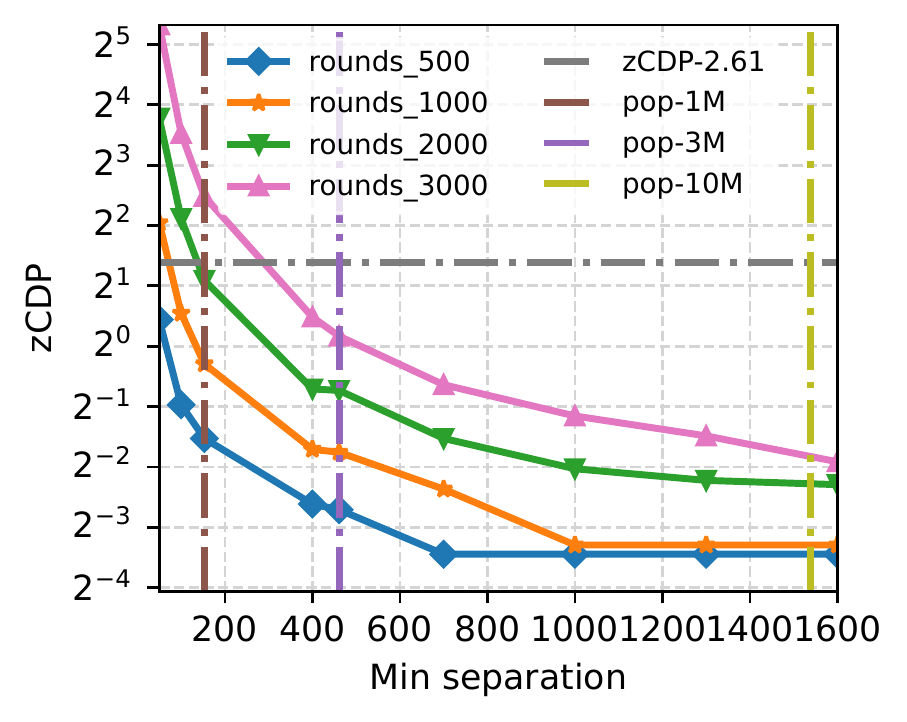}
\caption{Report goal 6500}
\label{fig:min_sep}
\end{subfigure}
\caption{The effect of population size, number of rounds, report goals, and min separation on DP-FTRL privacy guarantees. For a fixed number of rounds to achieve utility target, increasing report goal and min separation can achieve stronger guarantees measured by smaller zCDP. 
}\label{fig:privacy_computation}
\end{figure*}
}

\paragraph{Client participation.}
DP-FTRL achieves strong privacy if each client only participates once during training, or the number of client participation is limited when a client can participate multiple times. Two parameters are introduced to characterize client participation for DP-FTRL: the maximum participations (MaxP) of a client in all training rounds and the minimum round separation (MinS) between any single client's two participations.
MaxP and MinS are correlated as MaxP is upper bounded by rounds $T$ divided by MinS. In general, for fixed rounds $T$, decreasing MaxP and increasing MinS can lead to stronger privacy guarantees without changing utility. In addition, \citet{cho2023convergence} suggests potential advantage of increasing MinS for utility.

When using the worst-case MaxP estimated by rounds $T$ divided by MinS, \Cref{fig:min_sep} shows increasing MinS can achieve stronger privacy measured by smaller zCDP values. However, the maximum MinS is limited by the population size divided by the number of clients per round lower bounded by the report goal. For example, when the report goal is 6500 for small population of around $10^6$, MinS has to be smaller than $153$ rounds, so strong privacy guarantees are difficult to achieve when training for 3000 rounds. While we cannot measure the precise population size in the FL system due to client dynamics, we estimate the population size of various Gboard tasks as ranging from $0.8$ million to $16.6$ million in \cref{tab:nwp_fix_pretrain_largeswor}.

\paragraph{Report goal.} We study report goal for privacy-computation trade-off based on a hypothesis used in \citep{mcmahan18learning,kairouz21b,xu2022learning}: for sufficiently large data, the utility is approximately non-decreasing if the noise multiplier and clients per round (lower bounded by report goal) proportionally increase. 
We provide empirical justification to this hypothesis by comparing the evaluation accuracy of two training runs: one with a report goal of 500 and noise multiplier of 0.54, versus another of report goal 6500 and noise multiplier 7. On more than three Gboard language tasks, we observed that the final utility remains similar, or slightly better for larger report goals. Moreover, using a larger report goal speeds up learning at the beginning of training.
Based on the hypothesis, we plot \cref{fig:pop1m,fig:pop3m} by linearly increasing report goal and noise multiplier, and assuming the MinS is set to the maximum possible value (population divided by report goal) for strong privacy. Though a large report goal can limit the MinS, it generally leads to stronger privacy guarantees for reasonable population size and total rounds.

\paragraph{System configuration.} According to \cref{fig:pop1m,fig:pop3m}, we choose a large report goal of 6500 supported by the large scale FL systems and aim for maximum MinS for DP-FTRL. To control MinS in practice, a timer is introduced on clients in the FL system so that a client will only become eligible to participate in training (again) after a certain period of time has passed. \citet{dpftrl_blogpost} used a timer period of 24 hours to train the es-ES NWP model, which led to an observed MinS of 313. The MinS of es-ES is upper bounded by $4.21M/6500\sim647$ and can be potentially improved by increasing the timer period. We increase the timer period in the unit of 24 hours due to the uneven diurnal participation pattern \citep{yang2018applied,zhu2022diurnal}, and generally observe that MinS can proportionally increase with the timer period before reaching the possible maximum. However, there are many factors in the FL system that may affect the wall clock training speed, which makes it challenging to optimize the timer period to maximize MinS. 

\subsection{Public pretraining} \label{sec:pretrain}
 We explore pretraining on public data for production models, which were shown to substantially improve model utility in DP simulations~\citep{li2021large,de2022unlocking,yu2021differentially,xu2022learning,wang2023can}.
 We pretrain a model for each Gboard language task using the multi-lingual C4 dataset \citep{ColinRaffel2019ExploringTL,xue2020mt5} collected from public web pages. \Cref{fig:engb_acc} shows that pretraining can reduce $\sim 1000$ rounds to reach a given utility threshold under the same noise multiplier, which can significantly improve the privacy guarantees as shown in \cref{fig:privacy_computation}.
 
 We additionally observe that:
\begin{enumerate*}[label=\color{purple}(\arabic*)]
     \item it is  challenging to fine-tune from a pretrained model when the word embeddings are shared for
     input and output to reduce the parameter size of LMs for on-device deployment;
     \item the accuracy may decrease in the first a few rounds of fine-tuning;
     \item pretraining helps with diminishing marginal returns: at some point further pretraining does not necessarily improve the final performance.
\end{enumerate*}
Therefore, we use models with separate input and output embeddings and pretrain with half of the C4 dataset for Gboard LMs.

\subsection{Combining with secure aggregation} \label{sec:secagg}

Secure aggregation (SecAgg)~\cite{bonawitz2017practical} is a cryptographic multiparty computation protocol ensures that the central server can only access the aggregated update from a large set of clients, preventing inspection of individual client updates. We combine SecAgg and DP-FTRL to provide strong data minimization and anonymization protection \citep{bonawitz2021federated}. This work considers central DP and honest-but-curious server, and not the setting where the DP mechanism is applied distributively (i.e. on the client) as in \cite{kairouz2021distributed, agarwal2021skellam}. We describe the algorithm in \cref{algo:secagg-dp-ftrl}, and provide detailed discussion tackling the main challenge: how we can properly calibrate the sensitivity for DP when using SecAgg.

\begin{algorithm*}[bth]
\caption{Federated DP-FTRL with \colorbox{pink!25}{SecAgg}}
\label{algo:secagg-dp-ftrl}
\begin{algorithmic}
\INPUT: report goal $m$, learning rate for model weights on client $\eta_{c}$ and on server $\eta_{s}$, momentum $\beta=0.9$, noise multiplier for model delta $z_\Delta$,  total number of rounds $T$, \colorbox{pink!25}{fixed clip norm $C$}, and \colorbox{pink!25}{a large scaling parameter $s$}.
\begin{multicols}{2}
\STATE Initialize model $\theta^0$, momentum buffer $\bar{\Delta}^0=0$
\STATE Initialize tree $\tree$ with $z_\Delta$ and $C$
\STATE \colorbox{pink!25}{$d \leftarrow \text{dimension of } \theta^0$}
\STATE \colorbox{pink!25}{$C_\infty \leftarrow \text{ceil} \left(2 s \cdot C \log(d) / \sqrt{d} \right) $}
\STATE \colorbox{pink!25}{$M \leftarrow 2 \cdot C_\infty \cdot m + 1  $ \COMMENT{SecAgg's modulus}}
\STATE \colorbox{pink!25}{$\alpha \leftarrow \exp\left(-0.5\right)  $}
\STATE \colorbox{pink!25}{$\tilde H_{d} \leftarrow \frac{1}{\sqrt{d}} H_{d}$ \COMMENT{Normalized Hadmard matrix}} 

\FOR{each round $t=0, 1, 2, \ldots, T$}
\STATE $\mathcal{Q}^t \leftarrow$ ($m$ users for this round)
\STATE \colorbox{pink!25}{$\xi \leftarrow$ uniform random sign vector $ \in \{-1,+1\}^d$}
\STATE \colorbox{pink!25}{$D_\xi \leftarrow$  diagonal matrix with $\xi$ on the diagonal}

\FOR{each user $i \in \mathcal{Q}^t$ \textbf{in parallel}}
\STATE $\Delta^{t}_i \leftarrow \text{ClientUpdate}(i, \theta^t, \xi)$
\ENDFOR

\STATE \colorbox{pink!25}{\COMMENT{Securely aggregate the model updates}}

 \STATE  \colorbox{pink!25}{$\Delta_t = \sum_{i \in \mathcal{Q}^k} \Delta^{t}_i \mod{M}$}
  
\STATE \colorbox{pink!25}{\COMMENT{Unshift, unscale, and unrotate}}
 \STATE  \colorbox{pink!25} {$\Delta_t \leftarrow (1/s) \cdot D_\xi \cdot \tilde{H}_d^T \cdot \left(\Delta_t -C_\infty \right)$}
  
\STATE \COMMENT{Update model weights with noise addition}

\STATE $\tilde{\Delta}^t = \frac{1}{m} \psum \left(\tree, \, {\Delta_t}, \, k \in [0, t] \right)$

\STATE $\bar{\Delta}^t = \beta \bar{\Delta}^{t-1} + \tilde{\Delta}^t$, $\theta^{t+1} \leftarrow \theta^0 + \eta_{s} \bar{\Delta}^t$ 

\ENDFOR

\vspace{1.5ex}
\begin{mdframed}[
    linecolor=black,
    linewidth=1.5pt,
    roundcorner=4pt,
    userdefinedwidth=1.15\linewidth,
]
\FUNCTION{ClientUpdate($i$, $\theta_0, \xi$)}
\STATE $\theta \leftarrow \theta_0$

\STATE \colorbox{pink!25}{$D_\xi \leftarrow$  diagonal matrix with $\xi$ on the diagonal}
\STATE $\mathcal{G} \leftarrow $ (user $i$'s local data split into batches)
\FOR{batch $g \in \mathcal{G}$}
\STATE $\theta \leftarrow \theta - \eta_c \nabla \ell(\theta;g)$
\ENDFOR
\STATE $\Delta \leftarrow \theta - \theta_0$

\STATE\colorbox{pink!25}{$\Delta \leftarrow \Delta \cdot s \cdot \min{\left(1, \frac{C}{||\Delta||_2}\right)} \quad$  \COMMENT{Scale and $\ell_2$ clip}}

\STATE\colorbox{pink!25}{$\Delta \leftarrow \tilde H_{d} \cdot D_\xi \cdot \Delta$  \COMMENT{Randomly rotate}}

\STATE\colorbox{pink!25}{$\Delta \leftarrow \Delta \cdot \min \left(1, \frac{C_{\infty}}{||\Delta||_{\infty}}  \right)$  \COMMENT{Clip to bound $\ell_\infty$}}

\colorbox{pink!25}{ \COMMENT Conditional stochastic rounding}
 \REPEAT
 \STATE $\Delta'\gets$ stochastically round the coordinates of  $\Delta$
  \UNTIL{$\scriptstyle \|\Delta'\|_2^2 \le  s^2C^2 + d/4 + \sqrt{2\log(1/\alpha)} \cdot \left(sC + \sqrt{d} / 2\right)$}
  
\colorbox{pink!25}{\COMMENT Shift to center around $C_\infty$ and apply modulo}
\STATE \colorbox{pink!25}{\textbf{return} $\Delta' + C_\infty  \mod M$}
\ENDFUNCTION
\end{mdframed}
\end{multicols}

\end{algorithmic}
\end{algorithm*}

\paragraph{SecAgg overview.} We start by overviewing how SecAgg works at a high level. For an in depth treatment of the cryptographic protocol, we refer the reader to \cite{bell2020secure,bonawitz2017practical}. We will treat SecAgg as a ``black box'' which is guaranteed to faithfully compute the modular sum of integer vectors in a finite group, while not revealing any information about any individual vector beyond what can be learned from the sum; our methods do not depend on the specifics of the implementation of SecAgg. Concretely, for a vector $\Delta_i \in \mathbb{Z}_M^d$ representing the client's model update after all its elements being appropriately represented in a finite group of size $M$ (i.e., $\Delta_i$ is a $d$-dimensional vector with integer entries in the finite set $\{0, \cdots, M\}$), using SecAgg as a ``black box'', the server obtains: 
\begin{equation}
\Delta := \sum_i^m \Delta_i \! \mod M,
\end{equation}
and applies the DP-FTRL protocol to $\Delta$.
SecAgg's finite group, $M$, dictates the number of bits needed per parameter, and $d \log_2 M$ bits are needed per model update vector. 

\paragraph{Sensitivity analysis.} Our goal is to show that if the model updates are clipped to an $\ell_2$ norm of $C$ prior to being represented as integer vectors, the sensitivity of $\Delta$ can be bounded and quantified, and then we can rely on the DP analysis of DP-FTRL to do privacy accounting. Before diving into the sensitivity analysis, we first overview the steps needed to transform a model update vector in $\mathbb{R}^d$ to one in $\mathbb{Z}_M^d$, which is what the SecAgg protocol requires. These steps are performed on the client device; they are shown in Algorithm \ref{algo:secagg-dp-ftrl} inside of the ClientUpdate function (and highlighted in pink).

The first step is to clip the model update to bound its $\ell_2$ norm to  $C$ and then scaling it by a factor $s$ (a large constant). Notice that this clipping operation is required for DP even if we did not want to use SecAgg---only the post-clipping scaling is new. As we will see later in the discretization step, scaling by $s$ will prove crucial in terms of minimizing the inflation in sensitivity. Next, we apply a uniform random rotation, which helps us go from an $\ell_2$ geometry to an $\ell_\infty$ geometry. Precisely, a random rotation preserves the $\ell_2$ norm and provides a "high probability" bound on the $\ell_\infty$ norm: for $x \in \mathbb{R}^d$, the coordinates of $U_{\text{rotate}}x$ are sub-Gaussians with $\sigma^2 = ||x||^2_2 / d$ (see Lemma 28 of \cite{kairouz2021distributed}). For an efficient implementation, we randomly rotate $x$ by first randomizing the signs of its coordinates and then multiplying it by an appropriately scaled Hadamard matrix (see Lemma 29 of \cite{kairouz2021distributed}). After this step, each coordinate of the clipped, scaled, and randomly rotated model update vector is in the range $[-2s C/\sqrt{d}, 2s C/\sqrt{d}]$ with high probability\footnote{The factor of $2$ is to ensure that the range we consider contains $4\sigma$. We could have used a larger factor, but it would require increasing $M$, the modulus of SecAgg. In practice, we observed that $2$ performs well.}. Applying a union bound, the $\ell_\infty$ norm of this vector is in the range  $[-2s C \log(d) /\sqrt{d}, 2s C \log(d)/\sqrt{d}]$ with high probability. We therefore apply another clipping operation to bound the $\ell_\infty$ of the vector to $C_\infty = \text{ceil} (2 s C \log(d) / \sqrt{d})$. This bounds the range of every coordinate in the vector while guaranteeing that: (a) its $\ell_2$ norm is not inflated; and (b) the signal is preserved since this clipping operation is expected to be a ``no op'' with high probability. We are finally ready to move from $\mathbb{R}^d$ to $\mathbb{Z}^d$. To this end, we leverage the \textit{conditional stochastic rounding} procedure introduced in~\cite{kairouz2021distributed} to obtain a bounded norm on the scaled and rounded client vector.

Let $\tilde x$ be a stochastic rounding of vector $x \in \mathbb R^d$ to the integer grid $\mathbb Z^d$. Then, for $\alpha \in (0, 1)$, we have 
\begin{equation}
\label{prop:bounded-rounded-norm}
\mathbb P\left[ \|\tilde{x}\|_2^2 \le \|x\|_{2}^{2}+ d / 4 + \sqrt{2 \log (1 / \alpha)} \cdot\left(\|x\|_{2} + \sqrt{d} / 2 \right) \right] \ge 1 - \alpha.
\end{equation}
Conditional stochastic rounding is thus defined as retrying the stochastic rounding on $x$ until $\|\tilde{x}\|^2_2$ is within the probabilistic bound above. We show, in Algorithm \ref{algo:secagg-dp-ftrl}, how we can specialize this approach to our context, yielding discretized model update vectors in $\{-C_\infty, \cdots, C_\infty\}^d$ with an $\ell^2_2$ norm smaller than $s^2C^2 + d/4 + \sqrt{2\log(1/\alpha)} \cdot \left(sC + \sqrt{d} / 2\right)$. We used $\alpha = \exp(-0.5)$ in production training. We finally shift the vectors so that they have coordinates centered around $C_\infty$, i.e. in $\{0, \cdots, +C_\infty, \cdots, 2C_\infty\}^d$, and apply modulo $M$ so that the vector entries are in the right format that is expected by SecAgg. 

After applying the SecAgg protocol, we unshift, unscale, and unrotate the securely aggregated client outputs. It's important to notice that $M = 2 C_\infty \cdot m + 1$ was deliberately chosen so that no modulo wrap-arounds (overflows) happen during the secure aggregation step\footnote{We did not explicitly use clipping to bound the $\ell_\infty$ of client vectors in the current version of training. This means that modulo wrap-arounds could have happened, inflating the $\ell_2$ sensitivity by up to $dM$. However, our choice of $M$ was relatively large, and the probability of a modulo wrap-around is less than $2 \times 10^{-27}$.}. Upon unscaling by $1/s$, the $\ell^2_2$ norm of the vectors securely summed becomes $C^2_{\text{inflated}}= C^2 + d/(4s^2) + \sqrt{2\log(1/\alpha)} \cdot \left(C/s + \sqrt{d} / (2s^2)\right)$. Thus, the larger the $s$ is, the closer the inflated $\ell_2$ is to $C$ (the original clip norm applied in $\mathbb{R}^d$). In reality, we cannot choose $s$ to be arbitrarily large because $M$, which dictates the communication and computational costs of SecAgg, scales linearly with $s$. For DP accounting purposes, we have used the DP-FTRL accounting code with $C_{\text{inflated}}$ instead of $C$ to account for SecAgg's role.

\paragraph{Challenges and future work.} The large report goal requirement for strong DP guarantees is challenging for SecAgg in practice, which requires a slightly different system configuration. The SecAgg training speeds we observe are still notably slower, and we leave for future work potential improvements such as compression for communication efficiency~\citep{chen2022fundamental}, new DP methods to reduce report goal~\citep{choquette2022multi}, and embedding compression to reduce round time~\citep{shu2017compressing}.

\subsection{Recommended strategies and practices} \label{sec:practices}
We summarize our strategy for training Gboard LMs with DP. 
\begin{enumerate*}[label=\color{purple}(\arabic*)]
  \item Pre-train the model on public datasets if possible.
  \item Choose the maximum noise multiplier that meets the utility target based on small report goal simulation experiments on public datasets that is similar to the production task. 
  \item Based on the target number of rounds and estimated population, linearly increase the report goal and noise multiplier to meet the privacy target, and choose a large report goal supported by the system. If the privacy target is unachievable, fix the report goal to maximum, and increase the noise multiplier to target on a model with suboptimal utility.
  \item Estimate the possible maximum MinS based on chosen report goal and estimated population, and configure the timer period to approach the MinS; use previous experience of model training speed if applicable. 
  \item If the hyperparameters (e.g., learning rates) are known from previous experiments or simulation on public datasets, apply DP-FTRL with adaptive clipping (\cref{algo:ada-ftrl}) without manual tuning to try meet the privacy and utility goals. Note that \cref{algo:ada-ftrl} needs to account the noise inflation and restart for privacy guarantees. 
  \item If \cref{algo:ada-ftrl} fails or stronger privacy and utility are desirable, we can run a few small report goal experiments with \cref{algo:ada-ftrl} that tune quantile $\gamma$ and server learning rate $\eta_{s}$, select the best learning rate, and fix the clip norm based on the estimation; and run DP-FTRL with large report goals. 
  \item SecAgg can be used for all experiments, and precise MaxP and MinS are computed by post-processing for privacy accounting.
\end{enumerate*}


%% file: sec_alg.tex
\begin{algorithm*}[bth]
\caption{Federated DP-FTRL with \colorbox{blue!25}{adaptive clipping}}
\label{algo:ada-ftrl}
\begin{algorithmic}
\INPUT: report goal $m$, learning rate for model weights on client $\eta_{c}$ and on server $\eta_{s}$, momentum $\beta=0.9$, noise multiplier for model delta $z_\Delta$,  total number of rounds $T$, restart rounds $\restart=\{128+1024i, i=0, 1,\ldots\}$, \colorbox{blue!25}{quantile based norm estimation $C^0$}, \colorbox{blue!25}{target quantile $\gamma=0.5$}, \colorbox{blue!25}{learning rate for norm $\eta_\gamma=0.2$}, \colorbox{blue!25}{noise stddev for clip estimation $\sigma_b=m/20$} 
\begin{multicols}{2}
\STATE Initialize model $\theta^0$, momentum buffer $\bar{\Delta}^0=0$, clip norm $C_\theta=C^0$
\STATE Initialize tree $\tree_\theta$ with $z_\Delta$, $C_\theta$, \colorbox{blue!25}{and $\tree_b$ with $\sigma_b$}
\FOR{each round $t=0, 1, 2, \ldots, T$}
\STATE $\mathcal{Q}^t \leftarrow$ (at least $m$ users for this round)
\FOR{each user $i \in \mathcal{Q}^t$ \textbf{in parallel}}
\STATE $(\Delta^{t}_i, b^{t}_i) \leftarrow \text{ClientUpdate}(i, \theta^t, C_\theta, C^{t})$
\ENDFOR
\STATE \COMMENT{Update model weights with noise addition}
\STATE $\tilde{\Delta}^t = \frac{1}{m} \psum \left(\tree_\theta, \, \sum_{i \in \mathcal{Q}^k} \Delta^{k}_i, \, k \in [0, t] \right)$
\STATE $\bar{\Delta}^t = \beta \bar{\Delta}^{t-1} + \tilde{\Delta}^t$, $\theta^{t+1} \leftarrow \theta^0 + \eta_{s} \bar{\Delta}^t$ 
\STATE \colorbox{blue!25}{\COMMENT{Estimate quantile-based norm}}
\STATE \colorbox{blue!25}{$\tilde{b}^t = \frac{1}{m} \psum \left(\tree_b, \, \sum_{i \in \mathcal{Q}^k} b^{k}_i, \, k \in [0, t] \right)$}
\STATE \colorbox{blue!25}{$C^{t+1} \leftarrow C^0 \cdot \exp{\left( - \eta_\gamma(\tilde{b}^t - t \gamma)\right)}$}
\STATE \COMMENT{Restart and \colorbox{blue!25}{adjust clip norm}}
\IF{$t \in \restart $}
\STATE \colorbox{blue!25}{$C_\theta \gets C^{t+1}$}
\STATE Restart tree $\tree_\theta$ \colorbox{blue!25}{and $\tree_b$} with updated $C_\theta$
\ENDIF
\ENDFOR

\vspace{1.5ex}
\begin{mdframed}[
    linecolor=black,
    linewidth=1.5pt,
    roundcorner=4pt,
    userdefinedwidth=1.15\linewidth,
]
\FUNCTION{ClientUpdate($i$, $\theta_0$, $C_\theta$, $C$)}
\STATE $\theta \leftarrow \theta_0$
\STATE $\mathcal{G} \leftarrow $ (user $i$'s local data split into batches)
\FOR{batch $g \in \mathcal{G}$}
\STATE $\theta \leftarrow \theta - \eta_c \nabla \ell(\theta;g)$
\ENDFOR
\STATE $\Delta \leftarrow \theta - \theta_0$
\STATE \colorbox{blue!25}{$b \leftarrow \mathbb{I}_{||\Delta|| \leq C}$}
\STATE $\Delta' \leftarrow \Delta \cdot \min{\left(1, \frac{C_\theta}{||\Delta||}\right)} \quad$  \COMMENT{Clipping}
\STATE \textbf{return} $(\Delta', b)$
\ENDFUNCTION
\end{mdframed}
\end{multicols}
\end{algorithmic}
\vspace{-0.2cm}
\end{algorithm*}

%% file: sec_exp.tex

\begin{table*}[tbh]
    \centering
    \begin{tabular}{|c|c|c|c|c|c|c|c|c|}
         \hline
         \multirow{2}{*}{NWP} & \multirow{2}{*}{Rounds} & \multicolumn{2}{c|}{Utility} & \multicolumn{2}{c|}{Privacy} & \multirow{2}{*}{\shortstack[l]{Est. \\ Pop. (M) }} & \multirow{2}{*}{BaseModel} \\
         \cline{3-6}
         & & PRate(+\%) & Acc(+\%) & MinS/MaxP/Timer & zCDP & & \\
         \hline
         de-DE  & 930 & 8.28 
                & 12.49 
                & 212 / 4/ 48h
                & 0.48 & 3.24 & \multirow{5}{*}{\shortstack[l]{N-gram}} \\
         \cline{1-7}
         en-GB & 980 & 3.26 & 7.72 & 226 / 4 / 72h & 0.48 & 2.38 &\\
         \cline{1-7}
         fr-FR & 1280 & 3.78 & 8.50 & 180 / 5 / 72h & 0.89 & 2.79 &\\
         \cline{1-7}
         it-IT  & 1620 & 3.98 
                & 9.86 
                & 303 / 5 / 72h & 0.71 & 3.32  &\\
         \cline{1-7}
         pt-PT & 530 & 3.99 & 7.82 & 54 / 8 / 48h & 1.86 & 0.83  &\\
         \hline
         es-ES & 1900 & 0.29 & 0.48 & 526 / 3 / 144h & 0.35 & \multirow{2}{*}{4.21} & \multirow{2}{*}{\shortstack[l]{zCDP 0.81}} \\
         es-ES* & 1750 & 0.32 
            & 0.56 
            & 349 / 4 / 144h & 0.52 & & \\
         \hline
         en-US & 2800 & -0.39 & 0.11 & 371 / 7 / 48h & 1.31 & \multirow{2}{*}{13} & \multirow{4}{*}{\shortstack[l]{No-DP NN}}\\
         en-US*  & 1360 & -0.30 
                & 0.15 
                & 622 / 2 / 144h & 0.25 & &  \\
        \cline{1-7}
         pt-BR & 3600 & 0.18 & 0.29 & 909 / 3 / 144h & 0.45 & 16.6 &\\
        \cline{1-7}
         en-IN & 1290 & 0.19 & 0.40 & 170 / 6 / 96h & 1.14 & 7.72 &\\
         \cline{1-7}
         es-MX & 1980 & -0.15 & 0.29 & 343 / 5 / 96h & 0.64 & 9.96 &\\
         \hline
         es-AR & 640 & 0.25 & 3.50 & 90 / 5 / 96h & 0.84 & 4.09 & Mix \\
         \hline
    \end{tabular}
    \caption{Live A/B tests of DP NWP models. Utility shows the improvement from previously deployed models; privacy shows the key parameters and corresponding device-level zCDP; all models are trained by DP-FTRL with report goal of 6500 and noise multiplier of 7; en-US*/es-ES* are trained with SecAgg in addition to DP; the base model in AR is a mix of N-gram and No-DP NN models. 
    }
    \label{tab:nwp_fix_pretrain_largeswor}
    \vspace{-0.3cm}
\end{table*}

\begin{table*}[tbh]
    \centering
    \begin{tabular}{|c|c|c|c|c|c|c|}
         \hline
         \multirow{2}{*}{OTF} & \multirow{2}{*}{Rounds} & \multicolumn{2}{c|}{Utility} & \multicolumn{3}{c|}{Privacy} \\
         \cline{3-7}
           & & WMR(-\%) & WPM(+\%)   & MinS/MaxP/Timer & zCDP & DP-$ \epsilon (\delta=10^{-10})$ \\
         \hline
         de-DE & 1170 & 1.01
            & 0.59   
            & 206 / 5 / 48h &  0.89 & 9.01  \\
         \hline
         en-GB  & 1220 & 1.99
            & 0.38   
            & 206 / 5 / 72h & 0.89 & 9.01  \\
         \hline
         es-ES & 1280 &  1.03 
            & 0.60   
            & 197 / 5 / 48h & 0.89 & 9.01  \\
         \hline
         fr-FR  & 1300 & 1.83 
            & 0.67  
            & 290 / 4 / 72h & 0.61 & 7.31 \\
         \hline
         it-IT  & 1360 & 1.39
            & 0.80   
            & 188 / 5 / 48h & 0.89 & 9.01  \\
         \hline
         ru-RU  & 870 
            & 0.72 
            & 0.34  
            & 327 / 3 / 48h & 0.32 & 5.13  \\
         \hline
         pt-PT  & 430 & 1.71 & 0.32  & 54 / 7 / 48h & 0.99 & 9.56 \\
         \hline
    \end{tabular}
    \caption{Live A/B tests of DP OTF models. Utility shows the WMR decrease and WPM increase; privacy shows the key parameters and corresponding zCDP bound; all models are trained with DP-FTRL with report goal of 6500 and noise multiplier of 7; estimated population for ru-RU is 6.63M and other tasks can be found in \cref{tab:nwp_fix_pretrain_largeswor}.}
    \label{tab:otf_largeswor_pretrain}
\end{table*}

\section{Deploying DP LMs} \label{sec:exp}

\paragraph{A/B test metrics.}
We introduce metrics in A/B test to measure the utility of Gboard LMs. 
\begin{enumerate*}[label=\color{purple}(\arabic*)]
\item \emph{Picked Rate (PRate)}: the ratio of picked candidates among the NWP predictions; or SC predictions when it is triggered. 
\item \emph{Accuracy (Acc)}: the ratio of candidates matching the final committed words among the NWP model predictions. 
\item \emph{Trigger Rate}: the ratio of words with SC triggered among all committed words, which is an important metric when PRate is fixed. 
\item \emph{Word Modified Ratio (WMR)}: the ratio of words being modified during typing or after committed; improvement is shown by reduction.
\item \emph{Word Per Minute (WPM)}: the number of committed words per minute. 
\end{enumerate*}

\paragraph{Privacy guarantees.} Same as \citep{dpftrl_blogpost}, the zero-one device neighboring relationship (\citep[definition 1.1]{kairouz21b}) is adopted for DP. For user's with a single device, device-level DP corresponds directly to user-level DP. Our privacy guarantee holds for all well-behaved clients during training, and we do not account for privacy cost of modest amount of hyperparameter tuning.
DP is measured by the zero-Concentrated DP (zCDP) \citep{bun2016concentrated} guarantee that has been used by US census bureau \citep{bureau2021disclosure}, and can be easily converted to $(\epsilon, \delta)$-DP. 
We use the privacy accounting in \citep[appendix D]{kairouz21b} implemented in Tensorflow Privacy \citep{tfp}, and follow the guidelines outlined in \citep[Sec. 5.3]{ponomareva2023dpfy} to report detailed narratives of privacy guarantees in \cref{sec:privacy-guarantees}.

\ifthenelse{\boolean{acl}}{}{
\paragraph{Implementation.}
We open sourced implementation of DP-FTRL in Tensorflow Privacy \citep{tfp} integrated with Tensorflow Federated \citep{tff} as a DP aggregator for federated learning. Conceptually, DP-FTRL adds noise to the summation of updates across rounds, i.e., \emph{PrivateSum} in \cref{algo:ada-ftrl}. Instead of tracking the noise and summation separately, PrivateSum is implemented to only track the noise and updates $\tilde{\theta}^{t-1}$ by adding the residual of noise between round $t$ and round $t-1$. This design makes it easy to integrate with various optimizer choices, for example, the use of momentum, which is important for utility. It also allows ephemeral access of model deltas without directly storing unnoised states. 
}

\paragraph{Experimental setup.} We \ifthenelse{\boolean{acl}}{use the implementation in \cref{sec:impl}, and}{} apply the strategy in \cref{sec:practices} to train Gboard LMs with DP. We present NWP results in \cref{tab:nwp_fix_pretrain_largeswor}, and OTF results in \cref{tab:otf_largeswor_pretrain}. As Smart Compose (SC) reuses NWP LMs, SC has the same DP guarantees as NWP models by the post-processing property \citep{dwork2014algorithmic}. Following es-ES NWP model in \citep{dpftrl_blogpost}, we choose noise multiplier 7 and report goal 6500 based on simulation in \citep{kairouz21b} on public StackOverflow dataset \citep{stackoverflow}. We pretrain the models on public datasets and configure the timer period to control client participation, separately for different tasks. We use DP-FTRL with adaptive clipping and small report goal 500 to tune server learning rate and estimate the clip norm. Interestingly, we observe the learning rate and clip norm to be consistent for various Gboard LMs, and tuning seems to be unnecessary. DP-FTRL with fixed clip and large report goal is used to run the final model for deployment. 

\paragraph{Result analysis.} All NWP and OTF models in \cref{tab:nwp_fix_pretrain_largeswor,tab:otf_largeswor_pretrain} are trained with stronger guarantees (smaller zCDP) compared to zCDP $>2.6$ used by US Census Bureau~\citep{bureau2021disclosure}. For five NWP models in Europe (DE, GB, FR, IT, PT), the DP NN models significantly improve the utility compared to previous N-gram models. On en-US, pt-BR and en-IN, DP NN models also achieve comparable, or slightly better utility compared to their non-private versions as the strong models. SecAgg is successfully applied to en-US and es-ES, and can achieve good privacy-utility trade-off with a smaller number of rounds, likely due to the system configuration that results in more clients per round. However, SecAgg is also notably slower. There is a general positive correlation between the estimated population size and privacy guarantees. 

However, only a few tasks approach the possible maximum MinS for strong privacy guarantees, which highlights the challenge of both estimating population and controlling client participation. Longer training rounds are often used for NWP (compared to OTF) as the non-private NN baselines are strong, and to improve the downstream SC performance. As an example, we train es-ES NWP for 1900 rounds with a pretrained model, while the previous models \citep{dpftrl_blogpost} is trained for 2000 rounds without pretraining. Our es-ES NWP model slightly improves the utility measured by PRate and Acc, and improves the zCDP bound from 0.81 to 0.35 due to the larger MinS by timer configuration. We highlight that our es-ES model at round 1240 already achieves similar NWP utility and a strong privacy guarantee, but the utility of SC keeps improving with training. Compared to the previous model in \citep{dpftrl_blogpost}, our model improves the SC trigger rate by 4.23\% at round 1240, and 9.51\% at round 1900.


%% file: sec_con.tex
\section{Concluding remarks}
We discuss our experience and summarize our strategy for training production Gboard LMs with FL and DP. We propose an algorithm applying adaptive clipping \citep{andrew2019differentially} in DP-FTRL \citep{kairouz21b} to reduce the hyperparamter tuning. We discuss the impact on privacy and utility of several important factors: the clip norm, report goal, client participation, and pre-training. Our study highlights the importance of system and algorithm co-design for differential privacy in practice, the challenges of tuning in FL systems, and opportunities to improve the scalability and stability of FL with DP and/or SecAgg. More than twenty LMs with formal DP guarantees are trained and launched to support Gboard NWP, SC, and OTF features, including en-US and es-ES NWP models additionally trained with SecAgg. Our experience demonstrates the possibility of training DP models for practical applications when a large scale system is available for large scale data. Therefore, Gboard is introducing and enforcing a new policy: DP has to be applied in all future training and launching of Gboard LMs. 


%% file: sec_ack.tex
\subsection*{Acknowledgement}
The authors would like to thank Stanislav Chiknavaryan,  Adria Gascon, Zachary Garrett, and Timon Van Overveldt for infrastructure configuration support; Swaroop Ramaswamy, Om Thakkar, Abhradeep Thakurta for early discussion on models and algorithms; Jeremy Gillula for internal review process; Xu Liu, Shumin Zhai, and Daniel Ramage for leadership support.

%% file: sec_app_dpg.tex
\section{Reporting privacy guarantees}\label{sec:privacy-guarantees}
This section clarifies the nuances of the reported DP guarantees following the guidelines outlined in \citep[Sec. 5.3]{ponomareva2023dpfy}

\begin{enumerate}
    \item \textbf{DP setting}. This a central DP guarantee where the service provider is trusted to correctly implement the mechanism.
    \item \textbf{Instantiating the DP Definition}
     \begin{enumerate}
        \item \textit{Data accesses covered}: The DP guarantee applies to all well-behaved clients\footnote{Clients that faithfully follow the algorithm including participation limits. Due to the design of the algorithm, a mis-behaved client does not adversely affect the DP guarantee of any well-behaved clients.} in a single training run. We do not account for hyperparameter tuning, or the selection of the final model checkpoint using evaluation metrics or A/B testing in our guarantees. Public multilingual C4 data~\citep{ColinRaffel2019ExploringTL,xue2020mt5} is used for pre-training. 
        \item \textit{Final mechanism output}: Only the final model checkpoint is released for production launches, however the mechanism’s output is technically the full sequence of privatized gradients, and so the guarantee also applies at this level, and hence all intermediate models are protected (including those sent to devices participating in federated learning). 
        \item \textit{Unit of privacy}. Device-level DP is considered, i.e., the notion of adjacency is with respect to arbitrary training datasets on each client device, and the device might have an arbitrarily large local dataset containing arbitrary training examples. For user's with a single device, this corresponds directly to user-level DP; for devices shared with multiple users, this provides a stronger notion of DP than user-level; for a user with multiple devices that happen to both participate in training the model, the notion is weaker, but group privacy can be used to obtain a user-level guarantee.
        \item \textit{Adjacency definition for ``neigbouring'' datasets}: We use the zero-out definition~\citep{kairouz21b}. This is a a special form of the add-or-remove definition, where neighboring data sets differ by addition/removal of a single client. In the absence of a client at any training step, we assume that the client's model update gets replaced with the all zeros vector. This assumption enforces a subtle modification to the traditional definition of the add/remove notion of DP which allows neighboring data sets to have the same number of records.
    \end{enumerate}
    \item \textbf{Privacy accounting details}
    \begin{enumerate}
        \item \textit{Type of accounting used}: Both $\rho-$zCDP~\citep{bun2016concentrated} accounting, and PLD accounting~\citep{pldlib} for $(\epsilon, \delta)-$DP are used.
        \item \textit{Accounting assumptions }: Each client only participates limited times during the training, and there are at least a min-separation number of rounds between two consecutive participation of a client, i.e., MaxP and MinS as discussed in \cref{sec:hyper}. Client participation is enforced by a timer on clients in the cross-device FL system.    
        \item \textit{The formal DP statement}: The launched Gboard LMs have $\rho-$zCDP range in (0.2, 2). We also transform zCDP to $(\epsilon, \delta)-$DP by PLD accounting~\citep{pldlib}: given $\delta=10^{-10}$, the smallest zCDP $\rho=0.25$ corresponds to DP $\epsilon=4.49$; the largest zCDP $\rho=1.86$ corresponds to DP $\epsilon=13.69$.
        \item \textit{Transparency and verifiability}: We open sourced our core implementation code in TensorFlow Federated and Tensorflow Privacy. Key portions of the cross-device FL system are also open sourced.
    \end{enumerate}
\end{enumerate}

%% file: main.bbl
\begin{thebibliography}{40}
\providecommand{\natexlab}[1]{#1}
\providecommand{\url}[1]{\texttt{#1}}
\expandafter\ifx\csname urlstyle\endcsname\relax
  \providecommand{\doi}[1]{doi: #1}\else
  \providecommand{\doi}{doi: \begingroup \urlstyle{rm}\Url}\fi

\bibitem[Abadi et~al.(2016)Abadi, Chu, Goodfellow, McMahan, Mironov, Talwar,
  and Zhang]{abadi2016deep}
Martin Abadi, Andy Chu, Ian Goodfellow, H~Brendan McMahan, Ilya Mironov, Kunal
  Talwar, and Li~Zhang.
\newblock Deep learning with differential privacy.
\newblock In \emph{Proceedings of the 2016 ACM SIGSAC conference on computer
  and communications security}, pages 308--318, 2016.

\bibitem[Agarwal et~al.(2021)Agarwal, Kairouz, and Liu]{agarwal2021skellam}
Naman Agarwal, Peter Kairouz, and Ziyu Liu.
\newblock The skellam mechanism for differentially private federated learning.
\newblock \emph{Advances in Neural Information Processing Systems},
  34:\penalty0 5052--5064, 2021.

\bibitem[Andrew et~al.(2021)Andrew, Thakkar, McMahan, and
  Ramaswamy]{andrew2019differentially}
Galen Andrew, Om~Thakkar, H~Brendan McMahan, and Swaroop Ramaswamy.
\newblock Differentially private learning with adaptive clipping.
\newblock \emph{Conference on Neural Information Processing Systems (NeurIPS)},
  2021.

\bibitem[Balle et~al.(2020)Balle, Kairouz, McMahan, Thakkar, and
  Guha~Thakurta]{balle2020privacy}
Borja Balle, Peter Kairouz, Brendan McMahan, Om~Thakkar, and Abhradeep
  Guha~Thakurta.
\newblock Privacy amplification via random check-ins.
\newblock \emph{Advances in Neural Information Processing Systems},
  33:\penalty0 4623--4634, 2020.

\bibitem[Bell et~al.(2020)Bell, Bonawitz, Gasc{\'o}n, Lepoint, and
  Raykova]{bell2020secure}
James~Henry Bell, Kallista~A Bonawitz, Adri{\`a} Gasc{\'o}n, Tancr{\`e}de
  Lepoint, and Mariana Raykova.
\newblock Secure single-server aggregation with (poly) logarithmic overhead.
\newblock In \emph{Proceedings of the 2020 ACM SIGSAC Conference on Computer
  and Communications Security}, pages 1253--1269, 2020.

\bibitem[Bonawitz et~al.(2021)Bonawitz, Kairouz, McMahan, and
  Ramage]{bonawitz2021federated}
Kallista Bonawitz, Peter Kairouz, Brendan McMahan, and Daniel Ramage.
\newblock Federated learning and privacy: Building privacy-preserving systems
  for machine learning and data science on decentralized data.
\newblock \emph{Queue}, 19\penalty0 (5):\penalty0 87--114, 2021.

\bibitem[Bonawitz et~al.(2017)Bonawitz, Ivanov, Kreuter, Marcedone, McMahan,
  Patel, Ramage, Segal, and Seth]{bonawitz2017practical}
Keith Bonawitz, Vladimir Ivanov, Ben Kreuter, Antonio Marcedone, H~Brendan
  McMahan, Sarvar Patel, Daniel Ramage, Aaron Segal, and Karn Seth.
\newblock Practical secure aggregation for privacy-preserving machine learning.
\newblock In \emph{proceedings of the 2017 ACM SIGSAC Conference on Computer
  and Communications Security}, pages 1175--1191, 2017.

\bibitem[Bonawitz et~al.(2019)Bonawitz, Eichner, Grieskamp, Huba, Ingerman,
  Ivanov, Kiddon, Kone{\v{c}}n{\`y}, Mazzocchi, McMahan,
  et~al.]{bonawitz2019towards}
Keith Bonawitz, Hubert Eichner, Wolfgang Grieskamp, Dzmitry Huba, Alex
  Ingerman, Vladimir Ivanov, Chloe Kiddon, Jakub Kone{\v{c}}n{\`y}, Stefano
  Mazzocchi, Brendan McMahan, et~al.
\newblock Towards federated learning at scale: System design.
\newblock \emph{Proceedings of machine learning and systems}, 1:\penalty0
  374--388, 2019.

\bibitem[Bun and Steinke(2016)]{bun2016concentrated}
Mark Bun and Thomas Steinke.
\newblock Concentrated differential privacy: Simplifications, extensions, and
  lower bounds.
\newblock In \emph{Theory of Cryptography Conference}, pages 635--658.
  Springer, 2016.

\bibitem[Carlini et~al.(2019)Carlini, Liu, Erlingsson, Kos, and
  Song]{carlini2019secret}
Nicholas Carlini, Chang Liu, {\'U}lfar Erlingsson, Jernej Kos, and Dawn Song.
\newblock The secret sharer: Evaluating and testing unintended memorization in
  neural networks.
\newblock In \emph{28th USENIX Security Symposium (USENIX Security 19)}, pages
  267--284, 2019.

\bibitem[Chen et~al.(2022)Chen, Choo, Kairouz, and Suresh]{chen2022fundamental}
Wei-Ning Chen, Christopher A~Choquette Choo, Peter Kairouz, and Ananda~Theertha
  Suresh.
\newblock The fundamental price of secure aggregation in differentially private
  federated learning.
\newblock In \emph{International Conference on Machine Learning}, pages
  3056--3089. PMLR, 2022.

\bibitem[Cho et~al.(2023)Cho, Sharma, Joshi, Xu, Kale, and
  Zhang]{cho2023convergence}
Yae~Jee Cho, Pranay Sharma, Gauri Joshi, Zheng Xu, Satyen Kale, and Tong Zhang.
\newblock On the convergence of federated averaging with cyclic client
  participation.
\newblock \emph{arXiv preprint arXiv:2302.03109}, 2023.

\bibitem[Choquette-Choo et~al.(2022)Choquette-Choo, McMahan, Rush, and
  Thakurta]{choquette2022multi}
Christopher~A Choquette-Choo, H~Brendan McMahan, Keith Rush, and Abhradeep
  Thakurta.
\newblock Multi-epoch matrix factorization mechanisms for private machine
  learning.
\newblock \emph{arXiv preprint arXiv:2211.06530}, 2022.

\bibitem[De et~al.(2022)De, Berrada, Hayes, Smith, and Balle]{de2022unlocking}
Soham De, Leonard Berrada, Jamie Hayes, Samuel~L Smith, and Borja Balle.
\newblock Unlocking high-accuracy differentially private image classification
  through scale.
\newblock \emph{arXiv preprint arXiv:2204.13650}, 2022.

\bibitem[{DP Team}(2022)]{pldlib}
{DP Team}.
\newblock Google's differential privacy libraries., 2022.
\newblock \url{https://github.com/google/differential-privacy}.

\bibitem[Dwork et~al.(2014)Dwork, Roth, et~al.]{dwork2014algorithmic}
Cynthia Dwork, Aaron Roth, et~al.
\newblock The algorithmic foundations of differential privacy.
\newblock \emph{Foundations and Trends{\textregistered} in Theoretical Computer
  Science}, 9\penalty0 (3--4):\penalty0 211--407, 2014.

\bibitem[Geyer et~al.(2017)Geyer, Klein, and Nabi]{geyer2017differentially}
Robin~C Geyer, Tassilo Klein, and Moin Nabi.
\newblock Differentially private federated learning: A client level
  perspective.
\newblock \emph{arXiv preprint arXiv:1712.07557}, 2017.

\bibitem[Hard et~al.(2018)Hard, Rao, Mathews, Ramaswamy, Beaufays, Augenstein,
  Eichner, Kiddon, and Ramage]{hard2018gboard}
Andrew Hard, Kanishka Rao, Rajiv Mathews, Swaroop Ramaswamy, Fran{\c{c}}oise
  Beaufays, Sean Augenstein, Hubert Eichner, Chlo{\'e} Kiddon, and Daniel
  Ramage.
\newblock Federated learning for mobile keyboard prediction.
\newblock \emph{arXiv preprint arXiv:1811.03604}, 2018.

\bibitem[Kairouz et~al.(2019)Kairouz, McMahan, Avent, Bellet, Bennis, Bhagoji,
  Bonawitz, Charles, Cormode, Cummings, D'Oliveira, Rouayheb, Evans, Gardner,
  Garrett, Gasc{\'{o}}n, Ghazi, Gibbons, Gruteser, Harchaoui, He, He, Huo,
  Hutchinson, Hsu, Jaggi, Javidi, Joshi, Khodak, Konecn{\'{y}}, Korolova,
  Koushanfar, Koyejo, Lepoint, Liu, Mittal, Mohri, Nock, {\"{O}}zg{\"{u}}r,
  Pagh, Raykova, Qi, Ramage, Raskar, Song, Song, Stich, Sun, Suresh,
  Tram{\`{e}}r, Vepakomma, Wang, Xiong, Xu, Yang, Yu, Yu, and
  Zhao]{kairouz2019advances}
Peter Kairouz, H.~Brendan McMahan, Brendan Avent, Aur{\'{e}}lien Bellet, Mehdi
  Bennis, Arjun~Nitin Bhagoji, Kaylee Bonawitz, Zachary Charles, Graham
  Cormode, Rachel Cummings, Rafael G.~L. D'Oliveira, Salim~El Rouayheb, David
  Evans, Josh Gardner, Zachary Garrett, Adri{\`{a}} Gasc{\'{o}}n, Badih Ghazi,
  Phillip~B. Gibbons, Marco Gruteser, Za{\"{\i}}d Harchaoui, Chaoyang He, Lie
  He, Zhouyuan Huo, Ben Hutchinson, Justin Hsu, Martin Jaggi, Tara Javidi,
  Gauri Joshi, Mikhail Khodak, Jakub Konecn{\'{y}}, Aleksandra Korolova,
  Farinaz Koushanfar, Sanmi Koyejo, Tancr{\`{e}}de Lepoint, Yang Liu, Prateek
  Mittal, Mehryar Mohri, Richard Nock, Ayfer {\"{O}}zg{\"{u}}r, Rasmus Pagh,
  Mariana Raykova, Hang Qi, Daniel Ramage, Ramesh Raskar, Dawn Song, Weikang
  Song, Sebastian~U. Stich, Ziteng Sun, Ananda~Theertha Suresh, Florian
  Tram{\`{e}}r, Praneeth Vepakomma, Jianyu Wang, Li~Xiong, Zheng Xu, Qiang
  Yang, Felix~X. Yu, Han Yu, and Sen Zhao.
\newblock Advances and open problems in federated learning.
\newblock \emph{CoRR}, abs/1912.04977, 2019.
\newblock URL \url{http://arxiv.org/abs/1912.04977}.

\bibitem[Kairouz et~al.(2021{\natexlab{a}})Kairouz, Liu, and
  Steinke]{kairouz2021distributed}
Peter Kairouz, Ziyu Liu, and Thomas Steinke.
\newblock The distributed discrete gaussian mechanism for federated learning
  with secure aggregation.
\newblock In \emph{International Conference on Machine Learning}, pages
  5201--5212. PMLR, 2021{\natexlab{a}}.

\bibitem[Kairouz et~al.(2021{\natexlab{b}})Kairouz, Mcmahan, Song, Thakkar,
  Thakurta, and Xu]{kairouz21b}
Peter Kairouz, Brendan Mcmahan, Shuang Song, Om~Thakkar, Abhradeep Thakurta,
  and Zheng Xu.
\newblock Practical and private (deep) learning without sampling or shuffling.
\newblock In \emph{International Conference on Machine Learning (ICML)}, pages
  5213--5225, 2021{\natexlab{b}}.

\bibitem[Li et~al.(2021)Li, Tramer, Liang, and Hashimoto]{li2021large}
Xuechen Li, Florian Tramer, Percy Liang, and Tatsunori Hashimoto.
\newblock Large language models can be strong differentially private learners.
\newblock \emph{arXiv preprint arXiv:2110.05679}, 2021.

\bibitem[McMahan and Thakurta(2022)]{dpftrl_blogpost}
Brendan McMahan and Abhradeep Thakurta.
\newblock Federated learning with formal differential privacy guarantees, 2022.

\bibitem[McMahan et~al.(2017)McMahan, Moore, Ramage, Hampson, and
  y~Arcas]{mcmahan2017fedavg}
Brendan McMahan, Eider Moore, Daniel Ramage, Seth Hampson, and Blaise~Aguera
  y~Arcas.
\newblock Communication-efficient learning of deep networks from decentralized
  data.
\newblock In \emph{AISTATS}, pages 1273--1282. PMLR, 2017.

\bibitem[McMahan et~al.(2018)McMahan, Ramage, Talwar, and
  Zhang]{mcmahan18learning}
Brendan McMahan, Daniel Ramage, Kunal Talwar, and Li~Zhang.
\newblock Learning differentially private recurrent language models.
\newblock In \emph{International Conference on Learning Representations
  (ICLR)}, 2018.

\bibitem[Ponomareva et~al.(2023)Ponomareva, Hazimeh, Kurakin, Xu, Denison,
  McMahan, Vassilvitskii, Chien, and Thakurta]{ponomareva2023dpfy}
Natalia Ponomareva, Hussein Hazimeh, Alex Kurakin, Zheng Xu, Carson Denison,
  H.~Brendan McMahan, Sergei Vassilvitskii, Steve Chien, and Abhradeep
  Thakurta.
\newblock How to dp-fy ml: A practical guide to machine learning with
  differential privacy, 2023.

\bibitem[Raffel et~al.(2019)Raffel, Shazeer, Roberts, Lee, Narang, Matena,
  Zhou, Li, and Liu]{ColinRaffel2019ExploringTL}
Colin Raffel, Noam Shazeer, Adam Roberts, Katherine Lee, Sharan Narang, Michael
  Matena, Yanqi Zhou, Wei Li, and Peter~J. Liu.
\newblock Exploring the limits of transfer learning with a unified text-to-text
  transformer.
\newblock \emph{Journal of Machine Learning Research}, 2019.

\bibitem[Ramaswamy et~al.(2020)Ramaswamy, Thakkar, Mathews, Andrew, McMahan,
  and Beaufays]{ramaswamy2020training}
Swaroop Ramaswamy, Om~Thakkar, Rajiv Mathews, Galen Andrew, H~Brendan McMahan,
  and Fran{\c{c}}oise Beaufays.
\newblock Training production language models without memorizing user data.
\newblock \emph{arXiv preprint arXiv:2009.10031}, 2020.

\bibitem[Shu and Nakayama(2017)]{shu2017compressing}
Raphael Shu and Hideki Nakayama.
\newblock Compressing word embeddings via deep compositional code learning.
\newblock \emph{arXiv preprint arXiv:1711.01068}, 2017.

\bibitem[{TFF Authors}(2022{\natexlab{a}})]{stackoverflow}
{TFF Authors}.
\newblock Tensor{F}low {F}ederated {StackOverflow} dataset, 2022{\natexlab{a}}.
\newblock
  \url{https://www.tensorflow.org/federated/api\_docs/python/tff/simulation/datasets/stackoverflow}.

\bibitem[{TFF Authors}(2022{\natexlab{b}})]{tff}
{TFF Authors}.
\newblock Tensor{F}low {F}ederated., 2022{\natexlab{b}}.
\newblock \url{https://github.com/tensorflow/federated}.

\bibitem[{TFP Authors}(2022)]{tfp}
{TFP Authors}.
\newblock Tensor{F}low {P}rivacy., 2022.
\newblock \url{https://github.com/tensorflow/privacy}.

\bibitem[{US Census Bureau}(2021)]{bureau2021disclosure}
{US Census Bureau}.
\newblock Disclosure avoidance for the 2020 census: An introduction, 2021.

\bibitem[Wang et~al.(2023)Wang, Zhang, Cao, Li, McMahan, Oh, Xu, and
  Zaheer]{wang2023can}
Boxin Wang, Yibo~Jacky Zhang, Yuan Cao, Bo~Li, H~Brendan McMahan, Sewoong Oh,
  Zheng Xu, and Manzil Zaheer.
\newblock Can public large language models help private cross-device federated
  learning?
\newblock \emph{arXiv preprint arXiv:2305.12132}, 2023.

\bibitem[Wang et~al.(2021)Wang, Charles, Xu, Joshi, McMahan, Aguera~y Arcas,
  Al-Shedivat, Andrew, Avestimehr, Daly, et~al.]{wang2021fieldguide}
Jianyu Wang, Zachary Charles, Zheng Xu, Gauri Joshi, H~Brendan McMahan, Blaise
  Aguera~y Arcas, Maruan Al-Shedivat, Galen Andrew, Salman Avestimehr,
  Katharine Daly, et~al.
\newblock A field guide to federated optimization.
\newblock \emph{arXiv:2107.06917}, 2021.

\bibitem[Xu et~al.(2022)Xu, Collins, Wang, Panait, Oh, Augenstein, Liu,
  Schroff, and McMahan]{xu2022learning}
Zheng Xu, Maxwell Collins, Yuxiao Wang, Liviu Panait, Sewoong Oh, Sean
  Augenstein, Ting Liu, Florian Schroff, and H~Brendan McMahan.
\newblock Learning to generate image embeddings with user-level differential
  privacy.
\newblock \emph{arXiv preprint arXiv:2211.10844}, 2022.

\bibitem[Xue et~al.(2020)Xue, Constant, Roberts, Kale, Al-Rfou, Siddhant,
  Barua, and Raffel]{xue2020mt5}
Linting Xue, Noah Constant, Adam Roberts, Mihir Kale, Rami Al-Rfou, Aditya
  Siddhant, Aditya Barua, and Colin Raffel.
\newblock mt5: A massively multilingual pre-trained text-to-text transformer.
\newblock \emph{arXiv preprint arXiv:2010.11934}, 2020.

\bibitem[Yang et~al.(2018)Yang, Andrew, Eichner, Sun, Li, Kong, Ramage, and
  Beaufays]{yang2018applied}
Timothy Yang, Galen Andrew, Hubert Eichner, Haicheng Sun, Wei Li, Nicholas
  Kong, Daniel Ramage, and Fran{\c{c}}oise Beaufays.
\newblock Applied federated learning: Improving google keyboard query
  suggestions.
\newblock \emph{arXiv preprint arXiv:1812.02903}, 2018.

\bibitem[Yu et~al.(2021)Yu, Naik, Backurs, Gopi, Inan, Kamath, Kulkarni, Lee,
  Manoel, Wutschitz, et~al.]{yu2021differentially}
Da~Yu, Saurabh Naik, Arturs Backurs, Sivakanth Gopi, Huseyin~A Inan, Gautam
  Kamath, Janardhan Kulkarni, Yin~Tat Lee, Andre Manoel, Lukas Wutschitz,
  et~al.
\newblock Differentially private fine-tuning of language models.
\newblock \emph{arXiv preprint arXiv:2110.06500}, 2021.

\bibitem[Zhu et~al.(2022)Zhu, Xu, Chen, Kone{\v{c}}n{\`y}, Hard, and
  Goldstein]{zhu2022diurnal}
Chen Zhu, Zheng Xu, Mingqing Chen, Jakub Kone{\v{c}}n{\`y}, Andrew Hard, and
  Tom Goldstein.
\newblock Diurnal or nocturnal? federated learning of multi-branch networks
  from periodically shifting distributions.
\newblock In \emph{International Conference on Learning Representations}, 2022.

\end{thebibliography}
